\documentclass[journal,10pt,twocolumn]{IEEEtran}

\usepackage{amsmath}
\usepackage{upgreek}
\usepackage{bbold}
\usepackage{amssymb}
\usepackage{listings}
\usepackage{graphicx}
\usepackage{dsfont}
\usepackage{cite}
\usepackage{colortbl}
\usepackage{subcaption}
\usepackage{float}
\usepackage{amsmath,amssymb}
\usepackage{algorithm}
\usepackage{algorithmic}
\usepackage{comment}
\usepackage{xcolor}
\usepackage{hyperref}
\usepackage{amsthm}

\renewcommand{\a}{\boldsymbol{a}}

\renewcommand{\c}{\boldsymbol{c}}
\renewcommand{\v}{\boldsymbol{v}}
\renewcommand{\u}{\boldsymbol{u}}
\newcommand{\z}{\boldsymbol{z}}
\newcommand{\Z}{\boldsymbol{Z}}
\newcommand{\x}{\boldsymbol{x}}
\newcommand{\w}{\boldsymbol{w}}
\newcommand{\h}{\boldsymbol{h}}

\newcommand{\bSigma}{\boldsymbol{\Sigma}}
\newcommand{\bLambda}{\boldsymbol{\Lambda}}

\newcommand{\W}{\boldsymbol{W}}

\newcommand{\Y}{\boldsymbol{Y}}
\newcommand{\bT}{\boldsymbol{T}}
\newcommand{\R}{\boldsymbol{R}}
\newcommand{\F}{\boldsymbol{F}}
\newcommand{\D}{\boldsymbol{D}}
\newcommand{\C}{\boldsymbol{C}}
\newcommand{\B}{\boldsymbol{B}}
\newcommand{\A}{\boldsymbol{A}}
\newcommand{\I}{\boldsymbol{I}}
\renewcommand{\P}{\boldsymbol{P}}
\newcommand{\Q}{\boldsymbol{Q}}
\newcommand{\U}{\boldsymbol{U}}

\newcommand{\y}{\boldsymbol{y}}

\newcommand{\e}{\boldsymbol{e}}
\newcommand{\E}{\mathbb{E}}

\newcommand{\T}{\mathsf{T}}
\newcommand{\Tr}{\mathsf{Tr}}

\newtheorem{example}{Example}
\newtheorem{remark}{Remark}

\ifCLASSINFOpdf

\else

\fi

\begin{document}

\title{Efficient Image Splicing Localization via Contrastive Feature Extraction}

\author{Ronald~Salloum and C.-C.~Jay~Kuo
\thanks{The authors are with the Ming Hsieh Department of Electrical and Computer Engineering, University of Southern California, Los Angeles, CA, 90089 USA. Email: \{rsalloum, jckuo\}@usc.edu.}

}

\maketitle

\begin{abstract}
In this work, we propose a new data visualization and clustering technique for discovering discriminative 
structures in high-dimensional data. This technique, referred to as cPCA++, utilizes the fact that the interesting features of a ``target'' dataset may be obscured by high variance components during traditional PCA. By analyzing what is referred to as a ``background'' dataset (i.e., one that exhibits the high
variance principal components but not the interesting structures), our technique is capable of efficiently highlighting the structure that is unique to the ``target'' dataset. Similar to another recently proposed algorithm called ``contrastive PCA'' (cPCA), the proposed cPCA++ method identifies important dataset-specific patterns that are not detected by traditional PCA in a wide variety of settings. However, the proposed cPCA++ method is significantly more efficient than cPCA, because it does not require the parameter sweep in the latter approach. We applied the cPCA++ method to the problem of image splicing localization. In this application, we utilize authentic edges as the background dataset and the spliced edges as the target dataset. The proposed method is significantly more efficient than state-of-the-art methods, as the former does not require iterative updates of filter weights via stochastic gradient descent and backpropagation, nor the training of a classifier. Furthermore, the cPCA++ method is shown to provide performance scores comparable to the state-of-the-art Multi-task Fully Convolutional Network (MFCN).
\end{abstract}

\begin{IEEEkeywords}
Image Splicing, Multimedia Forensics, PCA, Contrastive PCA
\end{IEEEkeywords}

\IEEEpeerreviewmaketitle
\allowdisplaybreaks

\section{Introduction}

\begin{figure}
\centering
\includegraphics[width=1\columnwidth]{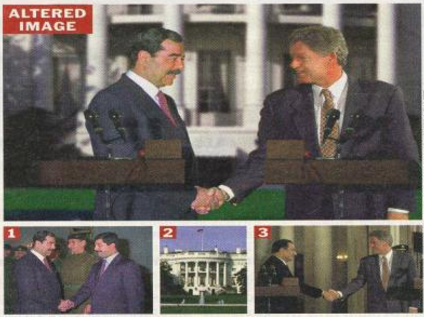}
\caption{An image splicing example. The manipulated image (in the top row) shows Bill Clinton shaking hands with Saddam Hussein. The three authentic images used to create the manipulated image are shown in the bottom row.}
\label{fig:clinton_example}
\end{figure}

\IEEEPARstart{W}{ith} the proliferation of low-cost and highly sophisticated image editing software, the generation of realistic-looking manipulated images has never been more easily accessible. At the same time, the prevalence of social media applications, such as Twitter, has made it very easy to quickly circulate these manipulated or fake images. Thus, there has been an increasing interest in developing forensic techniques to detect and localize forgeries (also referred to as manipulations or attacks) in images. One of the most common types of forgery is the image splicing attack. A splicing attack is a forgery in which a region from one image (i.e., the donor image) is copied and pasted onto another image (i.e., the host image). Forgers often use splicing to give a false impression that there is an additional object present in the image, or to remove an object from the image. Image splicing can be potentially used in generating false propaganda for political purposes. For example, the altered or manipulated image in the top row of Figure \ref{fig:clinton_example} shows Bill Clinton shaking hands with Saddam Hussein, although this event never occurred. The authentic images used to create the fake image can be seen in the bottom row of Figure \ref{fig:clinton_example}.

\begin{figure}
\centering
\includegraphics[width=1\columnwidth]{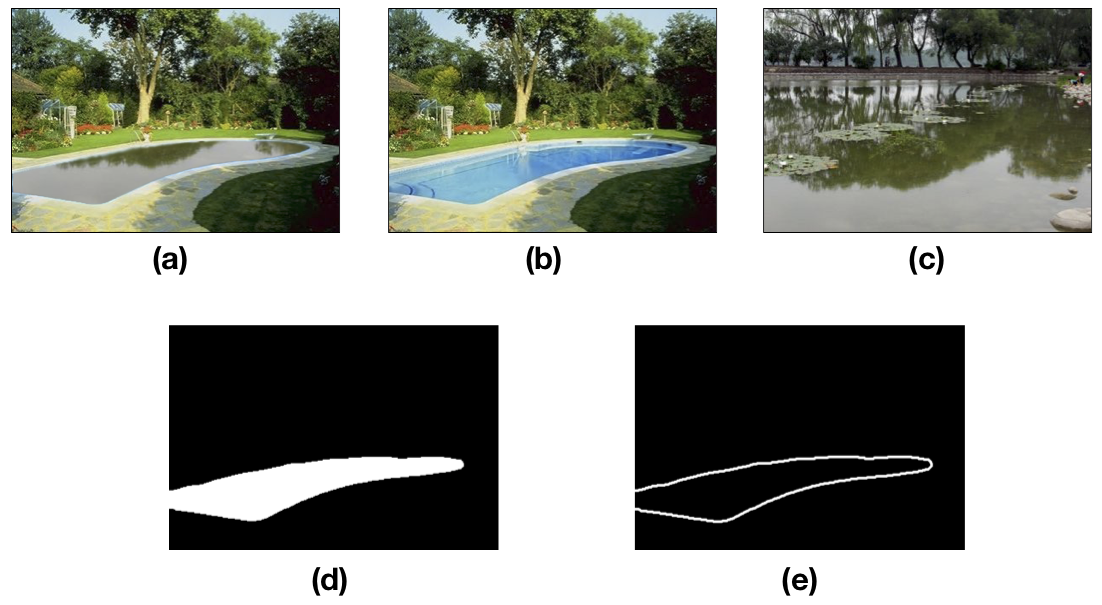}
\caption{An image splicing example showing: (a) the manipulated or probe image, (b) the host image, (c) the donor image, (d) the spliced surface (or region) ground truth mask, and (e) the spliced boundary (or edge) ground truth mask.}
\label{fig:additional_splicing_example}
\end{figure}

An additional splicing example\footnote{https://www.nist.gov/itl/iad/mig/media-forensics-challenge} is shown in Figure \ref{fig:additional_splicing_example}. The top row shows (from left to right): the manipulated image (also referred to as the probe or spliced image), the host image, and the donor image. In this example, the murky water from the donor image was copied and pasted on top of the clear water in the host image. This gives the false appearance that the water in the pool is not clean. The murky water in the manipulated image is referred to as the spliced surface or region. The bottom row in Figure \ref{fig:additional_splicing_example} shows (from left to right): the spliced surface (or region) ground truth mask and the spliced boundary (or edge) ground truth mask. These two types of masks provide different ways of highlighting the splicing manipulation. The surface ground truth mask is a per-pixel binary mask which specifies whether each pixel in the given manipulated image is part of the spliced surface (or region). We use the color white to denote a pixel belonging to the spliced surface and the color black to denote a pixel not belonging to the spliced surface. The edge ground truth mask is a per-pixel binary mask which specifies whether each pixel in the given probe image is part of the boundary of the spliced surface. In this case, we use the color white to denote a pixel belonging to the spliced boundary and the color black to denote a pixel not belonging to the spliced boundary.

There are two main problems in the splicing literature: detection and localization. The detection problem refers to the problem of classifying an image as a whole as either spliced or authentic, without localizing the spliced region or boundary. Many of the current techniques only address the detection problem, and do not address the localization problem. However, recently, more effort has been made in the forensics community in addressing the localization problem, which aims to classify each pixel in an image as either belonging to the spliced surface/boundary or not. Most of the recent techniques utilize deep-learning-based techniques such as convolutional neural networks (CNNs) to localize splicing attacks. Although CNNs have yielded some improvement in the localization of splicing attacks, they rely on careful selection of hyperparameters, network architecture, and initial filter weights. Furthermore, CNNs require a very long training time because they utilize stochastic gradient descent and backpropagation to iteratively update the filter or kernel weights. It is desired to have an approach that relies more on fundamental mathematical theory, does not require a significant amount of experimental tuning, and does not require long training and testing times. 

In this paper, we propose such an approach, the basis of which is Principal Component Analysis (PCA). As we will see, traditional PCA is not adequate for the task of image splicing localization, and so we will first present a new dimensionality-reduction technique for performing discriminative feature extraction, referred to as cPCA++, which is inspired by a recently proposed algorithm called ``contrastive PCA'' (cPCA) \cite{CPCA}. We then propose a new approach for image splicing localization based on cPCA++, which is significantly more efficient than state-of-the-art techniques such as the Multi-task Fully Convolutional Network (MFCN) \cite{salloum2018image}, and still achieves comparable performance scores. Also, cPCA++ is derived via matrix factorization, which allows us to identify underlying bases and corresponding weightings in the data that can be used for denoising applications (see App.~\ref{app:denoising} for such an example). 

The rest of the paper is organized as follows. Sec. \ref{sec:related-work} reviews existing splicing localization techniques. Sec. \ref{sec:cPCA++} presents cPCA++ as a new dimensionality-reduction technique, discusses how it differs from traditional PCA, and presents simulation results. The general framework or pipeline of the proposed cPCA++ approach for splicing localization is presented in Sec. \ref{sec:image_splice_localization_setup}. Experimental results on splicing datasets are presented in Sec. \ref{sec:experimental-analysis}. Finally, concluding remarks are given in Sec. \ref{sec:conclusion}.

\section{Related Work}
\label{sec:related-work}

Zampoglou et al. \cite{Zampoglou2017} conducted a comprehensive review of non-deep-learning-based techniques for image splicing localization, and provided a comparison of their performance. These techniques can be roughly grouped into the following three categories based on the type of feature or artifact they exploit: noise patterns \cite{lyu2014exposing,mahdian2009using,chen2008determining,li2012color}, Color Filter Array (CFA) interpolation patterns \cite{dirik2009image,ferrara2012image}, and JPEG-related traces \cite{lin2009fast,bianchi2012detection,farid2009exposing,li2009passive,ye2007detecting,luo2007novel,amerini2014splicing,bianchi2011improved,wang2010tampered}. Recently, there has been an increasing interest in the application
of deep-learning-based techniques to general image forensics and splicing detection/localization \cite{salloum2018image,pomari2018image,bondi2017tampering,rao2016deep,huh2018fighting,zhou2018learning,shi2018image,cozzolino2018noiseprint,chen2018improved}. Specifically, convolutional neural networks (CNNs) have attracted a significant amount of attention in the forensics community, due to the promising results they have yielded on a variety of image-based tasks such as object recognition and semantic segmentation \cite{long2015fully,simonyan2014very,krizhevsky2012imagenet}. One of the state-of-the-art CNN-based techniques for image splicing localization is the Multi-task Fully Convolutional Network (MFCN), proposed in \cite{salloum2018image}. In this work, one branch is used to learn the spliced surface ground truth mask (or label), while the other branch is used to learn the spliced boundary/edge ground truth mask. In addition to the surface ground truth mask, the boundaries between inserted regions and their host background can be an important indicator of a manipulated area. It is shown in \cite{salloum2018image} that simultaneously training on the surface and edge labels can yield finer localization of the splicing manipulation, as compared to training only on the surface label. It is also important to note that the MFCN-based method and the proposed cPCA++ method both output an edge-based probability map (as we will see in Sec. \ref{sec:image_splice_localization_setup}), which allows for a direct comparison.

Although CNN-based approaches have yielded promising results in the field of image forensics, they rely on careful selection of hyperparameters, network architecture, and initial filter weights. The behavior of the resulting CNN classifier, after learning of the coupled feature extractor and classifier blocks, is difficult to explain mathematically. Furthermore, CNNs require a long training time since the filter weights need to be iteratively updated via stochastic gradient descent and backpropagation. It is desired to have an approach that relies more on fundamental mathematical theory, does not require a significant amount of experimental tuning, and does not require long training/testing times. 

In this paper, we propose such an approach, based on a new version of Principal Component Analysis (PCA). In the context of image splicing localization, the two classes of interest (i.e., spliced and authentic boundaries) are very similar in terms of their covariance matrices, and traditional PCA is not able to effectively discriminate between the two classes. Instead, we propose a new version of PCA, referred to as cPCA++, that is able to perform discriminative feature extraction when dealing with extremely similar classes. We then propose a new approach for image splicing localization based on cPCA++. The proposed approach is mathematically tractable and does not require a significant amount of experimental tuning. Unlike CNNs, the proposed approach does not require iterative updates of filter weights via stochastic gradient descent and backpropagation, and thus is much more efficient than CNN-based approaches. In addition, we will see that the cPCA++ approach does not require the training of a classifier (e.g., support vector machines or random forests), which greatly speeds up the method. Also, the proposed approach can be readily parallelized due to its lack of dependence on inherently serial methods such as stochastic gradient descent. In the next section, we will derive the cPCA++ method, and then apply it to the image splicing localization problem in Sec.~\ref{sec:image_splice_localization_setup}.

\section{cPCA++}
\label{sec:cPCA++}

\subsection{The cPCA++ Method}
As we will see in Sec. \ref{sec:image_splice_localization_setup}, one characteristic of the image splicing localization problem is the strong similarity between authentic and spliced edges, especially when spliced edges have been smoothed over with a low-pass filter. Thus, traditional Principal Component Analysis (PCA) is not able to effectively discriminate between spliced and authentic edges. In order to deal with this issue, we will first study the problem of dimensionality reduction for extremely similar classes. 
Doing so will allow us to develop a new and efficient dimensionality-reduction technique that we refer to as cPCA++, which is inspired by a recently proposed algorithm called ``contrastive PCA'' (cPCA) \cite{CPCA}. In order to derive the cPCA++ method, we shall first consider the traditional PCA method.

Suppose we are presented with a data matrix $\Z\in\mathbb{R}^{M\times N}$, where $M$ denotes the original feature dimension of the data and $N$ denotes the number of instances included in $\Z$. PCA first computes the empirical covariance matrix of $\Z$: 
\begin{align}
    \widehat{\bSigma} \triangleq \frac{1}{N} \Z \Z^\T
\end{align}
where we assumed that $\Z$ has zero-mean. Next, PCA would compute the subspace spanned by the $K$ top or leading eigenvectors of $\widehat{\bSigma}$ (i.e., those corresponding to the largest $K$ eigenvalues). The basis for this space would constitute the \emph{filters} used to process the input data:
\begin{align}
    \F_{\mathrm{PCA}} \triangleq \mathrm{evecs}_K\left(\widehat{\bSigma}\right)
\end{align}
where $K$ denotes the number of principal eigenvectors to return. Now, a low-dimensional version of the input data $\Z$ can be obtained as:
\begin{align}
    \Y_{\mathrm{PCA}} = \F_{\mathrm{PCA}}^\T \Z.
\end{align}
It is known that the PCA filters $\F_{\mathrm{PCA}}$ preserve the most energy in $\Z$ after the transformation. Observe that this property may not yield  \emph{separability} of classes in $\Z$. That is, if the data matrix $\Z$ is composed of multiple classes (e.g., spliced and authentic edges in our case), performing traditional PCA will not necessarily allow us to find a representation in which we can separate the classes. 

In this work, we will take a different approach, inspired by a recently proposed algorithm called the ``contrastive PCA'' (cPCA) method \cite{CPCA}, which obtains discriminative filters. This approach focuses on finding directions that yield large variations for one dataset, referred to as the ``target'' or ``foreground'' dataset, while simultaneously yielding small variations for another dataset, referred to as the ``background'' dataset. There are two problem setups that benefit from this scheme, and we examine them in the following remark.

\begin{remark}
In \cite{CPCA}, the problem is to discover structures that are unique to one dataset relative to another. There are two ways of utilizing the solution to this problem. They are described as follows.

\begin{description}
    \item[1)] Consider the example described more thoroughly in Sec.~\ref{sec:mnist}. We are provided a ``target'' or ``foreground'' dataset with digits superimposed atop of grass background images. Observe that the target dataset contains multiple classes (i.e., the two digits, $0$ and $1$). Examples of instances of this dataset are illustrated in Fig.~\ref{fig:mnist_target_samples}. We wish to learn filters that are tuned to the digits, as opposed to the relatively strong grass background images. In order to accomplish this task, consider having access to a ``background'' dataset that only contains instances of grass images (i.e., without digits). The task is to discover the structures that are unique to the target dataset and not present in the background dataset. If this is accomplished, the filters are hoped to be able to differentiate between the two digits in the ``target'' dataset.

    \item[2)] In this alternative setup, we would like to solve a binary classification problem with the caveat that the two classes are very similar, and are dominated by the structures that are present in both classes. This is the case in the image splicing localization problem, where one class is the spliced edges and the other class is the authentic edges. In this scenario, we consider one class (i.e., the spliced edges) to be the target dataset and the other class (i.e., the authentic edges) to be the background dataset. If we are able to learn filters that are tuned to the structures unique to the spliced edges, then these filters form a basis for an efficient classifier that is able to differentiate between spliced and authentic edges. We will see the form of this classifier in Sec.~\ref{sec:image_splice_localization_setup}.
\end{description}

\end{remark}

We will take a detection-based approach to dimensionality reduction. We will setup a detector whose task is to identify whether or not a presented data matrix contains the interesting or special structures present in the foreground dataset (observe that the data matrix may contain background instances as well). By following this route, we will see that the detector will tune itself to specifically look for these special structures in the presented data matrix. Fortunately, the detector will look for these structures by transforming the incoming data matrix with a set of \emph{linear} filters. This means that we may simply utilize the output of these filters as a low-dimensionality representation of the input data matrix, tuned specifically to the interesting structures.

To setup the problem mathematically, we assume that the data matrix $\Z_b\in\mathbb{R}^{M\times N_b}$ is collected in an independently and identically distributed (i.i.d.) manner from the background dataset and has the following distribution:
\begin{align}
    \Z_b \sim  \mathcal{N}(\boldsymbol{0}_{M\times N_b}, \bSigma_b),\quad[\Z_b\textrm{ from background dataset}] \label{eq:Z0}
\end{align}
where $\bSigma_b\in\mathbb{R}^{M\times M}$ denotes an unknown covariance matrix of the background dataset and $\Z\sim\mathcal{N}(\A,\bSigma)$ with $\Z,\A \in\mathbb{R}^{M\times N}$ and positive-definite $\bSigma\in \mathbb{R}^{M\times M}$ means that $\Z$ has the following probability density function (pdf):
\begin{align}
    p(\Z ; \A, \bSigma) \triangleq \frac{\mathrm{exp}\left(-\frac{1}{2}\Tr[\bSigma^{-1} (\Z-\A)(\Z-\A)^\T]\right)}{\sqrt{(2\pi)^{M N} |\bSigma|^N}}.
\end{align}
On the other hand, we assume that when the instances of $\Z_f\in\mathbb{R}^{M\times N_f}$ are sampled independently from the foreground dataset, $\Z_f$ has the following distribution:
\begin{align}
     \Z_f \sim  \mathcal{N}(\W\Y_{\!\!f}, \bSigma_f),\quad[\Z_f\textrm{ from foreground dataset}]\label{eq:Z1}
\end{align}
where the mean of $\Z_f$ can be factored as the product of a basis matrix $\W\in\mathbb{R}^{M\times K}$ and a low-dimensionality representation matrix $\Y_{\!\!f}\in\mathbb{R}^{K\times N_f}$. The inner-dimension $K$ represents the underlying rank of the data matrix $\Z_f$, which is generally assumed to be much smaller than the initial dimension $M$.

Now, it is assumed that we are presented with a data matrix $\Z\in\mathbb{R}^{M\times N}$ that could be drawn completely from the background dataset (we refer to this case as the null hypothesis $\mathcal{H}_0$) or contain instances from the background dataset and some instances from the foreground dataset (we refer to this case as hypothesis $\mathcal{H}_1$). That is, the data matrix $\Z$ under $\mathcal{H}_1$ can be written as:
\begin{align}
    \Z \triangleq \left[\begin{array}{cc} \Z_f & \Z_b\end{array}\right] \label{eq:Z_structure_H1}
\end{align}
where $\Z_f\in\mathbb{R}^{M\times N_f}$ and $\Z_b\in\mathbb{R}^{M\times N_b}$ denote data instances from the foreground and background datasets, respectively. When $\mathcal{H}_1$ is active, we assume that we know the value of $N_f$. Given \eqref{eq:Z0} and \eqref{eq:Z1} above, we have that the mean of $\Z$ under $\mathcal{H}_1$ can be written as:
\begin{align}
    \E\left[\Z\right] &= \left[\begin{array}{cc} \E[\Z_f] & \E[\Z_b]\end{array}\right] \nonumber\\
    &=\left[\begin{array}{cc} \W\Y_{\!\!f} & \boldsymbol{0}_{M\times N_b}\end{array}\right] \nonumber\\
    &= \W \Y_{\!\!f} \bT
\end{align}
where we have introduced the matrix $\bT\in\mathbb{R}^{N_f\times N}$, where $N\triangleq N_f+N_b \geq 2$, with the following structure:
\begin{align}
    \bT \triangleq \left[\begin{array}{cc} \I_{N_f\times N_f} & \boldsymbol{0}_{N_f\times N_b}\end{array}\right].
    \label{eq:T_structure}
\end{align}
Therefore, the detection/classification problem we wish to examine is the determination of the active hypothesis (either $\mathcal{H}_0$ or $\mathcal{H}_1$), where the data matrix is described, under each hypothesis, by:
\begin{align}
    \Z \sim \begin{cases}
            \mathcal{N}(\boldsymbol{0}_{M\times N}, \bSigma_0), & \mathcal{H}_0\\
            \mathcal{N}(\W \Y_{\!\!f} \bT, \bSigma_1), & \mathcal{H}_1\\
         \end{cases}.
         \label{eq:factoring_equation}
\end{align}
Note that $\bSigma_0$ and $\bSigma_1$ denote the covariance matrix under $\mathcal{H}_0$ and $\mathcal{H}_1$, respectively.
The underlying assumption is that when the data matrix contains only instances from the background dataset, then the mean would be zero. We thus assume that all data instances of $\Z$ have the mean of the respective partitions subtracted off. For example, if the raw (unprocessed) data matrix is given by $\widetilde{\Z} \triangleq  \left[\begin{array}{cc} \widetilde{\Z}_f & \widetilde{\Z}_b\end{array}\right]$, then the partitions of the data matrix $\Z =\left[\begin{array}{cc} \Z_f & \Z_b\end{array}\right]$ are given by:
\begin{align}
    \Z_b &\triangleq \widetilde{\Z_b} - \frac{1}{N_b} \widetilde{\Z}_b \mathds{1}_{N_b} \mathds{1}_{N_b}^\T \label{eq:centering_Z0}\\
    \Z_f &\triangleq \widetilde{\Z_f} - \frac{1}{N_f} \widetilde{\Z}_f \mathds{1}_{N_f} \mathds{1}_{N_f}^\T. \label{eq:centering_Z1}
\end{align}

In solving this detection problem, we will be able to obtain the matrix $\Y_{\!\!f}\in\mathbb{R}^{K\times N_f}$, which is the desired low-dimensionality  representation of  the  foreground dataset. While it may initially appear that the detection problem is relatively simple---especially when the mean under the foreground dataset is relatively large in magnitude compared to the power in the covariance matrices, the detection problem is difficult when the mean is small compared to the covariance power (i.e., it is masked by large variation that naturally occurs in both the background and foreground datasets), as outlined earlier and in \cite{CPCA}. This is true in the context of image splicing localization, so it is assumed that the mean under the foreground dataset is small compared to the covariance power.

Now, we would like to determine if a given data matrix $\Z$ belongs to the distribution in the null hypothesis ($\mathcal{H}_0$) or the alternative hypothesis ($\mathcal{H}_1$). A classical approach to such problems is the likelihood ratio test, which is derived via the test:
\begin{align}
    p(\mathcal{H}_1 | \Z) &\geq p(\mathcal{H}_0 | \Z) \nonumber\\
    p(\Z | \mathcal{H}_1) p(\mathcal{H}_1) &\stackrel{(a)}{\geq}  p(\Z | \mathcal{H}_0) p(\mathcal{H}_0)
\end{align}
where step $(a)$ is due to Bayes' rule, and $p(\mathcal{H}_0)$ and $p(\mathcal{H}_1)$ denote the prior probabilities of hypotheses $\mathcal{H}_0$ and $\mathcal{H}_1$, respectively. The above can readily be re-arranged as:
\begin{align}
    \frac{p(\Z | \mathcal{H}_1)}{p(\Z|\mathcal{H}_0)} \geq \gamma\triangleq \frac{p(\mathcal{H}_0)}{p(\mathcal{H}_1)}. \label{eq:LRT}
\end{align}
This is referred to as the ``likelihood ratio test.'' It evaluates the likelihood ratio of the data matrix $\Z$ under each hypothesis and compares it to a threshold given by $\gamma$. The likelihood ratio on the left-hand-side of \eqref{eq:LRT} is desired to have a large value when the hypothesis $\mathcal{H}_1$ is active, but is desired to take on a small value when the hypothesis $\mathcal{H}_0$ is active. The threshold on the right-hand-side of \eqref{eq:LRT} is generally swept to achieve different performance that trades the probability of detection to false-alarm. 
Observe that the likelihood ratio test is only useful when the probabilities $p(\Z | \mathcal{H}_1)$ and $p(\Z | \mathcal{H}_0)$ can be computed (i.e., the probability density function is completely known or all of its parameters can be estimated). In our case, however, we do not have knowledge of many parameters, including the mean components $\W$ and $\Y_{\!\!f}$, as well as the covariance matrices $\bSigma_0$ and $\bSigma_1$. In order to deal with this situation, a common approach is to utilize what is referred to as the generalized likelihood ratio test (GLRT), which optimizes over parameters that are unknown from within the ratio. In our case, the GLRT is given by \cite[p.~200]{kay_detection}:
\begin{align}
    \mathrm{GLRT}(\W) \triangleq \frac{\underset{\bSigma_1,\Y_{\!\!f}}{\max}\  f_1(\Z;\bSigma_1,\Y_{\!\!f})}{\underset{\bSigma_0}{\max}\ f_0(\Z;\bSigma_0)} \geq \gamma \label{eq:GLRT}
\end{align}
where $\gamma$ denotes the threshold parameter, $f_1(\Z;\bSigma_1,\Y_{\!\!f})$ denotes the Gaussian pdf of $\Z$ under the hypothesis $\mathcal{H}_1$, and  $f_0(\Z;\bSigma_0)$ denotes the Gaussian pdf of $\Z$ under the null-hypothesis $\mathcal{H}_0$. That is,
\begin{align}
    f_1(\Z;\bSigma_1,\Y_{\!\!f}) &= \frac{\exp\!\left(\!-\frac{1}{2}\!\Tr\!\left[\bSigma_1^{-1}(\!\Z\!-\!\W\Y_{\!\!f}\bT\!)(\!\Z\!-\!\W\Y_{\!\!f}\bT\!)^\T\right]\!\right)}{\sqrt{(2\pi)^{M N} |\bSigma_1|^N}}\\
    f_0(\Z;\bSigma_0) &= \frac{1}{\sqrt{(2\pi)^{M N} |\bSigma_0|^N}}
\exp\left(\!-\frac{1}{2}\Tr\left[\bSigma_0^{-1}\Z\Z^\T\right]\!\right).
\end{align}
In \eqref{eq:GLRT}, we clearly indicate that the resulting GLRT statistic will be a function of the matrix $\W$, which is the basis for the feature mean under $\mathcal{H}_1$. We will return to this point later in the derivation. Now, we can optimize each pdf over the unknown covariance matrix, independently, as prescribed by the GLRT in \eqref{eq:GLRT} to obtain that the optimal covariance matrices for each case is given by the empirical covariance matrix (i.e., the maximum-likelihood covariance matrix estimator for a Gaussian distribution \cite[p.~89]{Duda}), so that:
\begin{align}
    \max_{\bSigma_0}\  f_0(\Z;\bSigma_0) &= f_0\left(\Z;\frac{1}{N} \Z\Z^\T\right) \nonumber\\
    &= \left((2\pi e)^{M} \left|\frac{1}{N} \Z\Z^\T\right|\right)^{-N/2}
\end{align}
and the pdf for $\mathcal{H}_1$ is given by
\begin{align}
     \max_{\bSigma_1}&\  f_1(\!\Z;\bSigma_1,\!\Y_{\!\!f}\!) \nonumber\\
     &= f_1\left(\!\Z;\frac{1}{N} (\Z\!-\!\W\Y_{\!\!f}\bT)(\Z\!-\!\W\Y_{\!\!f}\bT)^\T\!,\! \Y_{\!\!f}\!\!\right) \nonumber\\
    &= \left((2\pi e)^{M} \left|\frac{1}{N} (\Z\!-\!\W\Y_{\!\!f}\bT)(\Z\!-\!\W\Y_{\!\!f}\bT)^\T\right|\right)^{-N/2}.
\end{align}
Thus, the GLRT simplifies to:
\begin{align}
    \mathrm{GLRT}&(\W) \nonumber\\
    &= \frac{\underset{\Y_{\!\!f}}{\max}\ \!\!\left((2\pi e)^{M}\! \!\left|\!\frac{1}{N} (\Z\!-\!\W\Y_{\!\!f}\bT)(\Z\!-\!\W\Y_{\!\!f}\bT)^\T\!\right|\!\right)^{-\frac{N}{2}}}{\left((2\pi e)^{M} \left|\frac{1}{N} \Z\Z^\T\right|\right)^{-\frac{N}{2}}} \nonumber\\
    &\stackrel{(a)}{\Longleftrightarrow} \frac{\left|\Z\Z^\T\right|}{\underset{\Y_{\!\!f}}{\min} \ \left|(\Z-\W\Y_{\!\!f}\bT)(\Z-\W\Y_{\!\!f}\bT)^\T\right|} \label{eq:GLRT_eqiv}
\end{align}
where step $(a)$ states that the GLRT statistic is equivalent to \eqref{eq:GLRT_eqiv}. Now, using the derivation in \cite{UtilizingWaveformFeatures}, we have that the GLRT statistic can be written as (see equation (8) and related derivations in App.~A from \cite{UtilizingWaveformFeatures}):
\begin{align}
    \mathrm{GLRT}(\W) &= \frac{\left|\W^\T (\Z_b \Z_b^\T)^{-1} \W\right|}{\left|\W^\T (\Z \Z^\T )^{-1} \W\right|}. \label{eq:GLRT_opt_Y}
\end{align}
We will now optimize over our free variable $\W$. When the null-hypothesis is active, it can be shown that the GLRT statistic is independent of $\W$. On the other hand, when $\mathcal{H}_1$ is active, we have that the GLRT is dependent on $\W$ and is given by:
\begin{align}
    \mathrm{GLRT}(\W) &\stackrel{(a)}{=} \frac{\left|\W^\T (\Z_b \Z_b^\T)^{-1} \W\right|}{\left|\W^\T (\Z_f \Z_f^\T + \Z_b \Z_b^\T)^{-1} \W\right|} \ [\mathcal{H}_1 \textrm{ is active}]\nonumber\\
    &= \frac{\left|\W^\T \R_b^{-1} \W\right|}{\left|\W^\T (N_f/N_b \cdot \R_f + \R_b)^{-1} \W\right|} \label{eq:GLRT_H1}
\end{align}
where $(a)$ is obtained via \eqref{eq:Z_structure_H1} and we have defined the following quantities:
\begin{align}
    \R_b \triangleq \frac{1}{N_b} \Z_b \Z_b^\T, \quad\quad \R_f \triangleq \frac{1}{N_f} \Z_f \Z_f^\T
\end{align}
where $\R_b$ denotes the second order statistic associated with the background dataset only, while $\R_f$ denotes the second order statistic associated with the foreground dataset only. Now, to ensure that the gap in the GLRT statistic between the two hypotheses is as large as possible, we must maximize the GLRT value over $\W$ when the hypothesis $\mathcal{H}_1$ is active (since the GLRT value is independent of $\W$ when the null-hypothesis is active). It turns out that the value of $\W$ that maximizes the GLRT is obtained by solving a generalized eigenvalue problem \cite{TraceRatio} \cite[pp.~454--455]{IntroStatPR}. The optimal choice for $\W$ turns out to be the principal eigenvectors of the matrix $\I_{M\times M}+N_f/N_b\cdot \R_f\R_b^{-1}$, or, equivalently\footnote{Assuming a square matrix $\C=\U \D \U^{-1}$ is diagonalizable with non-negative eigenvalues, the matrix $\I+\alpha \C$ for $\alpha > 0$ can be written as $\I+\alpha \C=\U\U^{-1}+\alpha\U \D \U^{-1} = \U(\I+\alpha \D)\U^{-1}$, which modifies all eigenvalues of $\C$ by adding one to a positive-scaled version of them. This does not change the order of the eigenvalues of $\C$ since they are assumed to be non-negative and thus $\I+\alpha \C$ and $\C$ share the same principal eigenvectors when $\alpha>0$ and the eigenvalues of $\C$ are non-negative.}, the principal eigenvectors of the matrix $\R_f\R_b^{-1}$ when the eigenvalues of $\R_f\R_b^{-1}$ are non-negative (we will show later that this is indeed the case\footnote{We will later show that the eigenvalues of $\R_b^{-1}\R_f$ are non-negative. Observe, however, that $\R_b^{-1}\R_f$ and $\R_f\R_b^{-1}$ are similar since $\R_b$ is non-singular \cite[p.~53]{HornJohnson}, which means that the eigenvalues of $\R_f\R_b^{-1}$ are also non-negative.}). While the solution to $\W$ itself is not directly interesting for the problem of image splicing localization (it is relevant for applications related to matrix factorization, denoising, and dictionary learning \cite{DictionaryLearning}---which we explore in App.~\ref{app:denoising}), the reduced-dimensionality matrix $\Y_{\!\!f}$ is much more relevant for dimensionality reduction purposes. 

The maximum-likelihood estimator for $\Y_{\!\!f}$ (which was used to obtain \eqref{eq:GLRT_opt_Y}) is written as 
\begin{align}
\Y_{\!\!f} = \F^\T \Z_f   \label{eq:Y_as_lin_comb_Z}
\end{align}
where $\F\in\mathbb{R}^{M\times K}$ is a bank of filters yet to be determined, but has the following dependence on $\W$ \cite{UtilizingWaveformFeatures}:
\begin{align}
    \F = \R_b^{-1} \W (\W^\T \R_b^{-1} \W)^{-1}. \label{eq:F(W)}
\end{align}
In fact, the filters in \eqref{eq:F(W)} can be shown to be the principal eigenvectors of the matrix:
\begin{align}
    \Q\triangleq \R_b^{-1}\R_f.
\end{align}
This can be seen by noting that the optimal $\W$ (being the eigenvectors of $\R_f\R_b^{-1}$) must satisfy $\R_f\R_b^{-1} \W = \W \D$ for some diagonal matrix $\D$ that contains the eigenvalues. From this, we have that $\R_b^{-1} \R_f\R_b^{-1} \W = \R_b^{-1} \W \D$. Denoting $\P=\R_b^{-1}\W \bLambda$ for some diagonal invertible scaling matrix $\bLambda$ that we are free to choose, we have that $\P$ denotes the eigenvectors of the matrix $\Q=\R_b^{-1}\R_f$ (since $\Q\P = \P\D$). Substituting this into \eqref{eq:F(W)}, we have that $\F=\P \bLambda^{-1} (\W^\T \R_b^{-1}\W)^{-1}=\P$, with the choice of the matrix $\bLambda=(\W^\T \R_b^{-1}\W)^{-1}$. Now, we would like to show that the matrix $\bLambda$ is diagonal. First, observe that $\bLambda = \bLambda \W^\T \R_b^{-1} \W \bLambda = \P^\T \R_b \P$. The matrix $\P^\T \R_b \P$ can be shown to be diagonal since the columns of $\P$ are $\R_b$-orthogonal for distinct eigenvalues and the columns of $\P$ that correspond to the same eigenvalue can be chosen to be $\R_b$-orthogonal \cite[p.~345]{SymmetricEvalProblem}. Thus, $\bLambda=\P^\T \R_b \P$ is a diagonal matrix. It can also be shown $\bLambda$ is positive definite (and thus invertible) as long as $\W$ has full column rank. Furthermore, the matrix $\W$ has full column rank because the matrix $\R_f\R_b^{-1}$ can be shown to be diagonalizable. This is because $\R_f\R_b^{-1}$ can be shown to be similar to $\R_b^{-1/2} \R_f \R_b^{-1/2}$ \cite[p.~81]{Laub}, which is real and symmetric and thus diagonalizable \cite[p.~97]{Laub}. 

The obtained filters attempt to examine the specific structures that differentiate samples from $\mathcal{H}_0$ and $\mathcal{H}_1$. This leads to a relatively straightforward and efficient method for dimensionality reduction, which is listed in Alg.~\ref{alg:cPCA++}. Also, note that it is assumed that the matrix $\R_b$ is positive-definite. If it is not positive-definite, it is customary to apply diagonal-loading (addition of $\rho \boldsymbol{I}_{M\times M}$ to the covariance matrix for a small $\rho>0$) to force the covariance matrix to be positive-definite when it is rank-deficient (although, if enough diverse data samples are available, it is rare that the covariance matrix is rank-deficient).
\begin{algorithm}
   \caption{cPCA++ Method}
   \label{alg:cPCA++}
\begin{algorithmic}
   \STATE {\bfseries Inputs:} background data matrix $\widetilde{\Z}_b \in \mathbb{R}^{M\times N_b}$; target/foreground data matrix: $\widetilde{\Z}_f \in \mathbb{R}^{M\times N_f}$; $K$: dimension of the output subspace
   \vspace{2mm}
   \begin{enumerate}
   \STATE Center the data $\widetilde{\Z}_b$, $\widetilde{\Z}_f$ by obtaining $\Z_b$ and $\Z_f$ via \eqref{eq:centering_Z0}--\eqref{eq:centering_Z1}
       \item Compute: 
       \vspace{-2mm}
       \begin{align}
           \R_b &= \frac{1}{N_b} \Z_b \Z_b^\T \\
           \R_f &= \frac{1}{N_f} \Z_f \Z_f^\T
       \end{align}
       \vspace{-3mm}
       \STATE Perform eigenvalue decomposition on 
       \begin{align}
           \Q = \R_b^{-1} \R_f \label{eq:Q_alg}
       \end{align}
       \vspace{-3mm}
       \STATE Compute the top $K$ right-eigenvectors $\F$ of $\Q$
  \end{enumerate}
  \STATE{\bfseries Return:} the subspace $\F \in\mathbb{R}^{M\times K}$
\end{algorithmic}
\end{algorithm}

It is not immediately clear that $\F$ is a real matrix, since $\Q$ may not be symmetric. However, since the matrix $\Q$ is similar to the real and symmetric matrix  $\R \triangleq \R_b^{-1/2} \R_f \R_b^{-1/2}$ \cite[p.~81]{Laub} (this is because $\Q$ is by definition similar to $\R_b^{1/2} \Q \R_b^{-1/2} = \R_b^{1/2} \R_b^{-1} \R_f \R_b^{-1/2} = \R_b^{-1/2} \R_f \R_b^{-1/2} = \R$) where $\R_b^{1/2}$ denotes the matrix square root of $\R_b$, we have that $\R$ and $\Q$ share a spectrum, which is real due to $\R$ being a real symmetric matrix \cite[p.~78]{Laub}. It is important to note that the eigenvalues of $\Q$ are also non-negative due to it being similar to $\R$, and $\R$ being a positive-semi-definite matrix\footnote{The definition of a positive semi-definite matrix $\R$ is that $\R$ is symmetric and satisfies $\v^\T \R \v \geq 0$ for all $\v\neq \mathbb{0}$. In our case, we have $\v^\T \R \v = \v^\T \R_b^{-1/2} \R_f \R_b^{-1/2} \v = \y^\T \R_f \y = \frac{1}{N_f} \y^\T \Z_f\Z_f^\T \y = \x^\T \x = \|\x\|^2\geq 0$, where $\y\triangleq \R_b^{-1/2} \v$ and we defined $\x\triangleq 1/\sqrt{N_f} \cdot \Z_f^\T \y$ as a transformation of $\y$.}. That is, since $\R$ is positive-semi-definite, its eigenvalues are non-negative, and since $\Q$ is similar to $\R$, its eigenvalues are also non-negative. This was a requirement for our earlier derivation to show that the principal eigenvectors of $\I_{M\times M}+N_f/N_b\cdot \R_f\R_b^{-1}$ are the same as the principal eigenvectors of $\R_f\R_b^{-1}$. We also know that the eigenvectors of the matrix $\R$ are real since $\R$ is real and symmetric \cite[p.~97]{Laub}. Let us now denote one of the eigenvectors of $\R$, associated with eigenvalue $\lambda\in\mathbb{R}$ as $\x\in\mathbb{R}^{M\times 1}$. Then, we have that $\R \x = \lambda \x$ and $\Q \v = \lambda \v$ for some eigenvector $\v\in\mathbb{C}^{M\times 1}$ of $\Q$ (since $\lambda$ is an eigenvalue of $\Q$ as we have shown above---we will see, however, that we can always find eigenvectors in $\mathbb{R}^{M\times 1}$, as opposed to $\mathbb{C}^{M\times 1}$). Multiplying both sides on the left by $\R_b^{-1/2}$, we obtain:
\begin{align}
    \R_b^{-1/2} \R \x &= \Q \R_b^{-1/2} \x  = \lambda \R_b^{-1/2} \x.
\end{align}
Then, we immediately obtain that:
\begin{align}
    \Q \v = \lambda \v
\end{align}
where 
\begin{align}
    \v &\triangleq \frac{\R_b^{-1/2} \x}{\|\R_b^{-1/2} \x\|_2} \in \mathbb{R}^{M\times 1}\label{eq:evec_Q}
\end{align} 
denotes the eigenvector of $\Q$, which is real due to $\x\in\mathbb{R}^{M\times 1}$ and $\R_b^{-1/2} \in \mathbb{R}^{M\times M}$. Thus, we see that $\F$ is indeed a real matrix. Note that $\F$ (which contains the principal $K$ eigenvectors of $\Q$ as its columns) may not satisfy $\F^\T\F = \I_{K\times K}$ since the eigenvectors of $\Q$ are not guaranteed to be orthonormal.

To see the advantage of the cPCA++ method, it is useful to compare it with the ``contrastive PCA'' (cPCA) algorithm from \cite{CPCA}. The cPCA algorithm from \cite{CPCA} performs an eigen-decomposition of the following matrix:
\begin{align}
    \Q_{\mathrm{cPCA}} \triangleq \R_f - \alpha \R_b
\end{align}
where $\alpha$ is a contrast parameter that is swept to achieve different dimensionality reduction output. We observe that the cPCA algorithm is inherently sensitive to the relative scales of the covariance matrices $\R_b$ and $\R_f$. For this reason, a parameter sweep for $\alpha$ (which adjusts the scaling between the two covariance matrices) was used in \cite{CPCA}. In other words, multiple eigen-decompositions are required in the cPCA algorithm. On the other hand, the cPCA++ algorithm does not require this parameter sweep, as the relative scale between the two covariance matrices will be resolved when performing the eigen-decomposition of the matrix $\Q$. To see this property analytically, consider Example \ref{ex:ex1}. 
\begin{example}[A needle in a haystack]
\label{ex:ex1}
We consider the following $M$-feature example in order to better understand the traditional PCA, cPCA, and cPCA++ methods, and how they react to a small amount of information hidden in a significant amount of ``noise.'' Let the covariance matrices of the background and target datasets be respectively given by (where we note that $\R_b$ is positive-definite):
\begin{align}
    \R_b &\triangleq \gamma \a \a^\T + \rho \I_{M\times M}\label{eq:example_neg_class_covar}\\
    \R_f &\triangleq \beta \a \a^\T + \epsilon \c \c^\T \label{eq:example_pos_class_covar}
\end{align}
where $\a,\c\in\mathbb{R}^{M\times 1}$ and it is assumed that $\|\a\|_2 = \|\c\|_2 = 1$ and $\a^\T \c = 0$, making the vectors $\a$ and $\c$ orthonormal. It is further assumed that $\epsilon \ll \beta$, which means that the vector that is not common to both background and target has very small power in comparison to what is common between them. It is also assumed that $\beta\cdot\rho/\gamma < \epsilon$ and $\rho\ll \gamma$, which means that the diagonal-loading should not affect the result significantly. It is assumed that $\gamma, \beta, \epsilon, \rho > 0$. In this example, we would like to obtain a filter to reduce the dimension to $K=1$ from our original $M$ features. Note that \eqref{eq:example_pos_class_covar} describes a matrix with eigenvalues $\beta$ and $\epsilon$ and associated eigenvectors $\a$ and $\c$. The reason why we are interested in the above problem is that by examining the covariance structure above, we see that the data is inherently noise-dominated. This noise (i.e., the $\a\a^\T$ term) appears in both the background dataset and the target dataset, but the target dataset may contain some class-specific information (i.e., the $\c\c^\T$ term) that is not present in the background dataset. However, this ``interesting'' information is dominated by the noise due to the eigenvector $\a$. We wish to perform dimensionality reduction that \emph{rejects} variance due to the eigenvector $\a$ but preserves information due to the eigenvector $\c$ of the target dataset. The obvious solution is to choose the dimensionality reduction filter to be $\c$. We will examine how each algorithm performs.
\begin{description}
    \item[\!\!\!\!\!\!\!\textbf{PCA}] When traditional PCA is executed on the target dataset, the principal eigenvector (and thus $\F_{\mathrm{PCA}}$) will be found to be $\a$ since $\epsilon \ll \beta$. In addition, since $\rho\ll \gamma$, $\a$ is also the principal eigenvector of the background dataset, which is not useful in extracting the interesting structure from the foreground dataset. This is because $\F_{\mathrm{PCA}}^\T \c = 0$, which means that this filter will in fact null-out the ``interesting'' eigenvector.
    \item[\!\!\!\!\!\!\!\textbf{cPCA}] When cPCA is executed on the background and target datasets, we have that $\Q_{\mathrm{cPCA}} = \R_f-\alpha\R_b$ will be given by:
    \begin{align}
        \Q_{\mathrm{cPCA}} &= (\beta-\alpha\gamma) \a \a^\T + \epsilon \c \c^\T - \alpha\rho \I_{M\times M}. \label{eq:example_cpca_Q}
    \end{align}
    Now, \eqref{eq:example_cpca_Q} allows us to see exactly the reasoning behind sweeping $\alpha$. When $\alpha=0$, we obtain the PCA method operating on the target dataset (which would pick the filter defined by $\a$). When $\alpha\rightarrow\infty$, we obtain the PCA method operating on the background dataset (which still would pick the filter defined by $\a$). Instead, the desired value of $\alpha$ is one that nulls out the eigenvector $\a$---e.g., $\alpha = \beta/\gamma$. Observe that when this choice of $\alpha$ is made (or chosen through a sweep), we have that the principal eigenvector (and thus $\F_{\mathrm{cPCA}}$) of $\Q_{\mathrm{cPCA}}$ will be $\c$. This is precisely what we want the filter to be in order to differentiate between the two datasets (or extract the useful information from the target dataset relative to the background). Although the optimal $\alpha$ value is evident in this simple analytic example (due to the fact that we know the underlying structure indicated in \eqref{eq:example_neg_class_covar} and \eqref{eq:example_pos_class_covar}), this is not the case in most experiments and thus it is necessary to sweep $\alpha$ over the unbounded range $\alpha\in[0,\infty)$.
    \item[\!\!\!\!\!\!\!\textbf{cPCA++}] For the analysis of the cPCA++ method, let us first consider the inverse of $\R_b$, via the matrix inversion lemma:
    \begin{align}
    \R_b^{-1} &= \frac{1}{\rho} \left(\I_{M\times M} - \frac{\gamma}{\rho + \gamma} \a\a^\T \right) \nonumber\\
    &\stackrel{(a)}{\approx} \frac{1}{\rho} \left(\I_{M\times M} - \a\a^\T \right) \nonumber\\
    &= \frac{1}{\rho} \P_{\a}^\perp 
    \end{align}
    where step $(a)$ is due to the assumption that $0 < \rho \ll \gamma$ and $\P_{\a}^\perp=\I_{M\times M} - \a\a^\T$ denotes the projection onto the left null-space of $\a$ \cite[p.~22]{Laub} (since $\P_{\a}^\perp \a = \mathbb{0}_M$). Now, multiplying $\R_b^{-1}$ by $\R_f$, we obtain:
    \begin{align}
        \Q &= \R_b^{-1} \R_f \nonumber\\
        &\approx \frac{1}{\rho}  \P_{\a}^\perp \left(\beta \a \a^\T + \epsilon \c \c^\T\right)\nonumber\\
        &= \frac{\epsilon}{\rho}  \P_{\a}^\perp  \c \c^\T \nonumber\\
        &\stackrel{(a)}{=} \frac{\epsilon}{\rho} \c \c^\T
    \end{align}
    where step $(a)$ is due to the fact that $\a$ and $\c$ are orthonormal. This means that the $\Q$ matrix in the case of the cPCA++ method is approximately a rank-$1$ matrix with principal eigenvector $\c$. Clearly then, the filter chosen by cPCA++ will be $\F = \c$, which is the desired result. Observe that this result was obtained without the need for a hyper-parameter sweep (i.e., like the sweep over the $\alpha$ parameter in cPCA).
    \end{description}
\end{example}

Note that another recent publication \cite{wang2018dpca}, also based on the general principle of \cite{CPCA}, was brought to our attention. However, careful review of this article shows that the approach in \cite{wang2018dpca} is different from ours. In \cite{wang2018dpca}, the algorithm was obtained in a similar manner as the cPCA algorithm (i.e., it was formulated as a slightly modified maximization problem). On the other hand, our proposed cPCA++ approach is fundamentally different because it directly addresses the matrix factorization and dimensionality reduction nature of our problem, and therefore, gives rise to an intuitive per-instance classification algorithm (see Alg.~\ref{alg:classify_patch} further ahead) that can be applied to image splicing localization, and also addresses image denoising problems (see App.~\ref{app:denoising}).

In the next subsection, we will simulate both the cPCA method and the cPCA++ method for a collection of experiments from \cite{CPCA} (we will also simulate the popular t-SNE dimensionality reduction method \cite{tsne}). We will observe that the cPCA++ algorithm will be able to discover features from within the target dataset without the need to perform a parameter sweep, which will improve the running time of the method significantly (see Sec.~\ref{sec:timing}).

\subsection{Performance Comparison of Feature Extraction Methods}
\label{sec:comparison_to_cPCA}
In this subsection, we will examine many of the same experiments that the cPCA algorithm\footnote{\url{https://github.com/abidlabs/contrastive}} was applied to in \cite{CPCA}. Table \ref{tbl:datasets_overview} lists the parameters of the various datasets that will be examined in this section. In all cases, the desired dimensions from the methods will be set to $K=2$ (i.e., after feature reduction). We also incorporated the t-Distributed Stochastic Neighbor Embedding (t-SNE) algorithm \cite{tsne} in the results\footnote{\url{https://lvdmaaten.github.io/tsne/##implementations}}, even though t-SNE does not provide an explicit function for the discovered feature reduction. For example, after learning the filters $\F$ through Alg.~\ref{alg:cPCA++}, it is possible to efficiently apply them to new data matrices. In the case of t-SNE, however, the dimensionality reduction is done in relationship to the other samples in the dataset, so when a new data matrix is to be reduced in dimensions, the t-SNE algorithm must be re-executed. However, we include it in this exposition since it is a common algorithm for modern feature reduction. The main take-away from these experiments is the fact that the cPCA++ algorithm can obtain an explicit filter matrix, $\F$, and does not require a parameter sweep, which makes the algorithm much more efficient. At the same time, it is still able to discover the interesting structures in the target dataset.

\begin{table}[!t]
\renewcommand{\arraystretch}{1.7}
\caption{Overview of the datasets used for comparing the PCA, cPCA, t-SNE, and cPCA++ dimensionality reduction methods. Note that $N_f$ denotes the number of foreground samples, $N_b$ denotes the number of background samples, and $M$ denotes the original feature dimension.}
\label{tbl:datasets_overview}
\centering
\begin{tabular}{c||c|c|c}
\hline \hline
\rowcolor[gray]{0.9} \rule[-1ex]{0pt}{4ex} \textbf{Example} & $N_f$ & $N_b$ & $M$\\ 
\hline
\rule[-1ex]{0pt}{4ex} \textbf{Synthetic} & $400$ & $400$ & $30$  \\ 
\hline 
\rule[-1ex]{0pt}{4ex} \textbf{MNIST over Grass} & $5000$ & $5000$ & $784$  \\ 
\hline 
\rule[-1ex]{0pt}{4ex} \textbf{Mice Protein Expression} & $270$ & $135$ & $77$  \\ \hline
\rule[-1ex]{0pt}{4ex} \textbf{MHealth Measurements} & $6451$ & $3072$ & $23$  \\
\hline
\rule[-1ex]{0pt}{4ex} \textbf{Single  Cell  RNA-Seq  of  Leukemia  Patient} & $7898$ & $1985$ & $500$  \\
\hline \hline
\end{tabular} 
\end{table}

\subsubsection{Synthetic Data}
\label{sec:synthetic_data}
We first consider a synthetic data example to help illustrate the differences between the traditional PCA, cPCA, and cPCA++ algorithms. Consider the following original feature structure (i.e., prior to feature reduction) for an arbitrary $i$-th sample from a foreground or target data matrix $\widetilde{\Z_f}$:
\begin{align}
    \tilde{\z}_{f,i}^k &= \left[\begin{array}{ccc}
         (\h_{i,1}^{k})^{\T} & (\h_{i,2}^{k})^{\T} & \h_{i,3}^\T
    \end{array}\right]^\T \label{eq:target_dataset_structure}
\end{align}
in which $k\in\{1,2,3,4\}$ denotes the class index, with sub-vectors $\h_{i,1}^k,\h_{i,2}^k,\h_{i,3}\in\mathbb{R}^{10\times 1}$, where $\h_{i,1}^k$, $\h_{i,2}^k$, $\h_{i,3}$ are independent. Now, it is assumed that $\h_{i,3} \sim \mathcal{N}(\mathbb{0}_{10},10\boldsymbol{I}_{10\times 10})$. On the other hand, the other two sub-vectors are more useful to differentiate across the different classes $k$:
\begin{align}
    \h_{i,1}^k \sim \begin{cases}
    \mathcal{N}(\mathbb{0}_{10}, \boldsymbol{I}_{10\times 10}), & k\in\{1,2\}\\
    \mathcal{N}(6\mathds{1}_{10}, \boldsymbol{I}_{10\times 10}), & k\in\{3,4\}
    \end{cases}
\end{align}
while $\h_{i,2}^k$ is distributed according:
\begin{align}
    \h_{i,2}^k \sim \begin{cases}
    \mathcal{N}(\mathbb{0}_{10}, \boldsymbol{I}_{10\times 10}), & k\in\{1,3\}\\
    \mathcal{N}(3\mathds{1}_{10}, \boldsymbol{I}_{10\times 10}), & k\in\{2,4\}.
    \end{cases}
\end{align}
Knowing the above structure, we can deduce that to differentiate between the different classes $1\leq k\leq 4$, we need only use $\h_{i,1}^k$ and $\h_{i,2}^k$, while completely ignoring $\h_{i,3}$ as it does not help differentiate between the different classes. On the other hand, traditional PCA executed on the target data matrix $\widetilde{\Z_f}$ would readily pick features from $\h_{i,3}$ since they have the largest variance---even though they do not help in distinguishing between the different classes. 

\begin{figure}[ht]
    \centering
    \includegraphics[width=1\columnwidth]{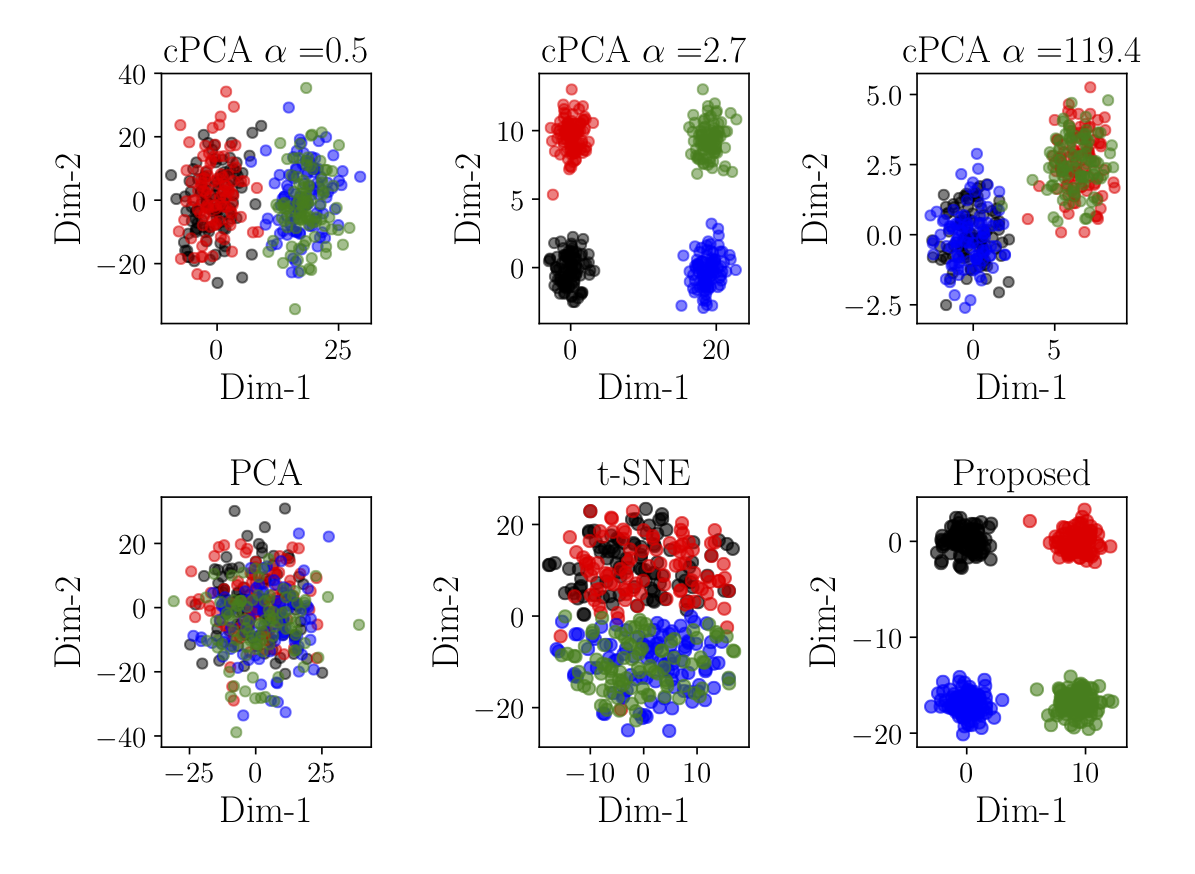}
    \caption{Performance on a synthetic dataset, where different colors are used for the four different classes. The top row of plots show the performance of the cPCA algorithm for different positive values of the contrast parameter $\alpha$. Clearly, a contrast factor for $\alpha=2.7$ is ideal, but must be found by sweeping $\alpha$. The bottom-left plot shows the performance of traditional PCA (which, as expected,  fails to separate the classes). The bottom-center plot shows the performance of the t-SNE algorithm, which again fails to discover the underlying structure in the high-dimensional data. Finally, the bottom-right figure shows the output obtained by the cPCA++ method, which obtains the ideal clustering without a parameter sweep.}
    \label{fig:cpca:synthetic}
\end{figure}

Instead, it is assumed that there is a background dataset that, while not containing the differentiating features, follows the following distribution:
\begin{align}
    \tilde{\z}_{b,i} \sim \left[\begin{array}{c}
     \mathcal{N}(\mathbb{0}_{10}, 3\boldsymbol{I}_{10\times 10}) \\
     \mathcal{N}(\mathbb{0}_{10}, \boldsymbol{I}_{10\times 10}) \\
     \mathcal{N}(\mathbb{0}_{10}, 10\boldsymbol{I}_{10\times 10})
    \end{array}\right] \label{eq:synthetic_bg}
\end{align}
where the sub-vectors are independent.
We observe that $\tilde{\z}_{b,i}$ only informs us as to the relative variance across the dimensions, but not the explicit mean values (i.e., feature structure) within the dimensions. It can be shown that the filters obtained by the cPCA++ algorithm are given by (please see App.~\ref{app:synthetic}):
\begin{align}
    \F &= \left[\begin{array}{cc}
    \e_1 & \e_2 
    \end{array}\right] \otimes \frac{1}{\sqrt{10}} \mathds{1}_{10} \label{eq:synthetic_filters}
\end{align}
where $\e_n$ denotes the standard basis vector of dimension $3$ (i.e., $\e_n = [\mathbb{0}_{n-1}^\T, 1, \mathbb{0}_{3-n}^\T]^\T$). Note that these are the ideal filters to extract the signal structure from the target dataset described by \eqref{eq:target_dataset_structure}. We ran the cPCA++ method on this synthetic dataset, and compared the results with the other feature reduction methods in Fig.~\ref{fig:cpca:synthetic}. Note that the t-SNE method operated over the target or foreground dataset. For all methods examined, we extract the dimensionality reduced dataset for $K=2$ dimensions. All classes in the figure are equally likely, and a total of $400$ samples were generated. In the top row of plots in Fig.~\ref{fig:cpca:synthetic}, we ran the cPCA algorithm for three positive choices of the contrast parameter $\alpha$. We see that a contrast factor of $\alpha=2.7$ yields a correct separation among the classes, while the other values of the contrast factor fail to do so.  In the bottom-left plot, we have the performance of the traditional PCA algorithm on the target dataset. It is in fact a special case of the cPCA method outlined in \cite{CPCA} for a contrast parameter of $\alpha=0$. We observe that traditional PCA fails to cluster the data. This is expected by our investigation above, since the traditional PCA method is designed to pick the directions that best explain the most variance of the dataset, which are directions that explain $\h_3$. In the bottom-center plot, we executed the t-SNE algorithm \cite{tsne} on the foreground dataset for $250$ iterations (no significant improvement was obtained by running t-SNE longer). We see that t-SNE also fails to select the correct dimensions with which to visualize the data. Finally, in the bottom-right plot, we show the data plotted against the feature directions obtained by the cPCA++ algorithm. We observe that our feature extractor was able to perform the correct feature selection in one step, without the need for a parameter sweep. We note that while this is a synthetic example that can be solved analytically, it helps in illustrating how the method actually works. In the next few simulations, we compare and contrast the performance of the cPCA++, cPCA, traditional PCA, and t-SNE algorithms on more complex datasets.

\subsubsection{MNIST over High Variance Backdrop}
\label{sec:mnist}
In this experiment (also conducted in \cite{CPCA}), the MNIST dataset \cite{MNIST} is used and the $28\times28$ images corresponding to digits $0$ and $1$ are extracted. Utilizing traditional PCA processing, it is possible to separate these two particular digits.

\begin{figure}[ht]
    \centering
    \includegraphics[width=1\columnwidth]{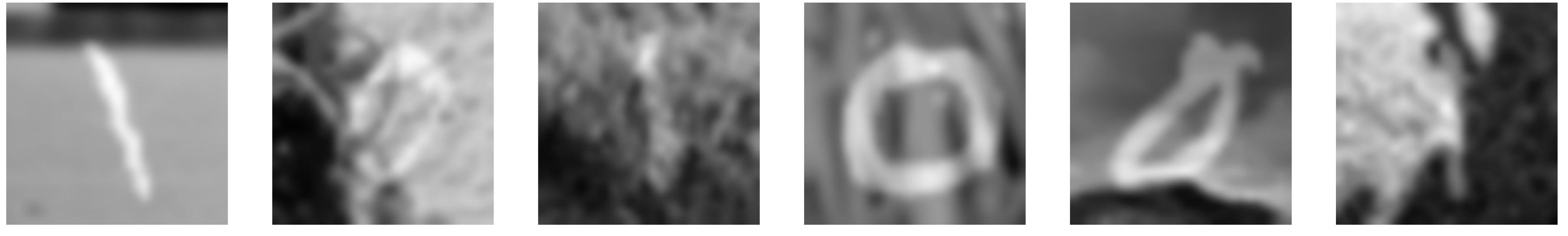}
    \caption{Example of six target images. The MNIST images for digits $0$ and $1$ are superimposed on top of grass images.}
    \label{fig:mnist_target_samples}
\end{figure}

However, the authors of \cite{CPCA} (as we do here), super-imposed these digits atop of high-variance grass backgrounds in order to mask the fine details that make the digit recognizable and distinct. Figure \ref{fig:mnist_target_samples} shows six examples of these ``target'' images. We see that the digits themselves are difficult to see atop the high-variance background images. The task now is to apply the feature reduction techniques to still separate these digits.
In order to obtain a ``background'' dataset for the cPCA and cPCA++ methods, grass images (without the MNIST digits) randomly sampled from the ImageNet dataset \cite{Imagenet} were used to estimate $\R_b$. Note that the same grass images that were used to generate the target images need not be utilized, but grass images in general will do. 

\begin{figure}[ht]
    \centering
    \includegraphics[width=1\columnwidth]{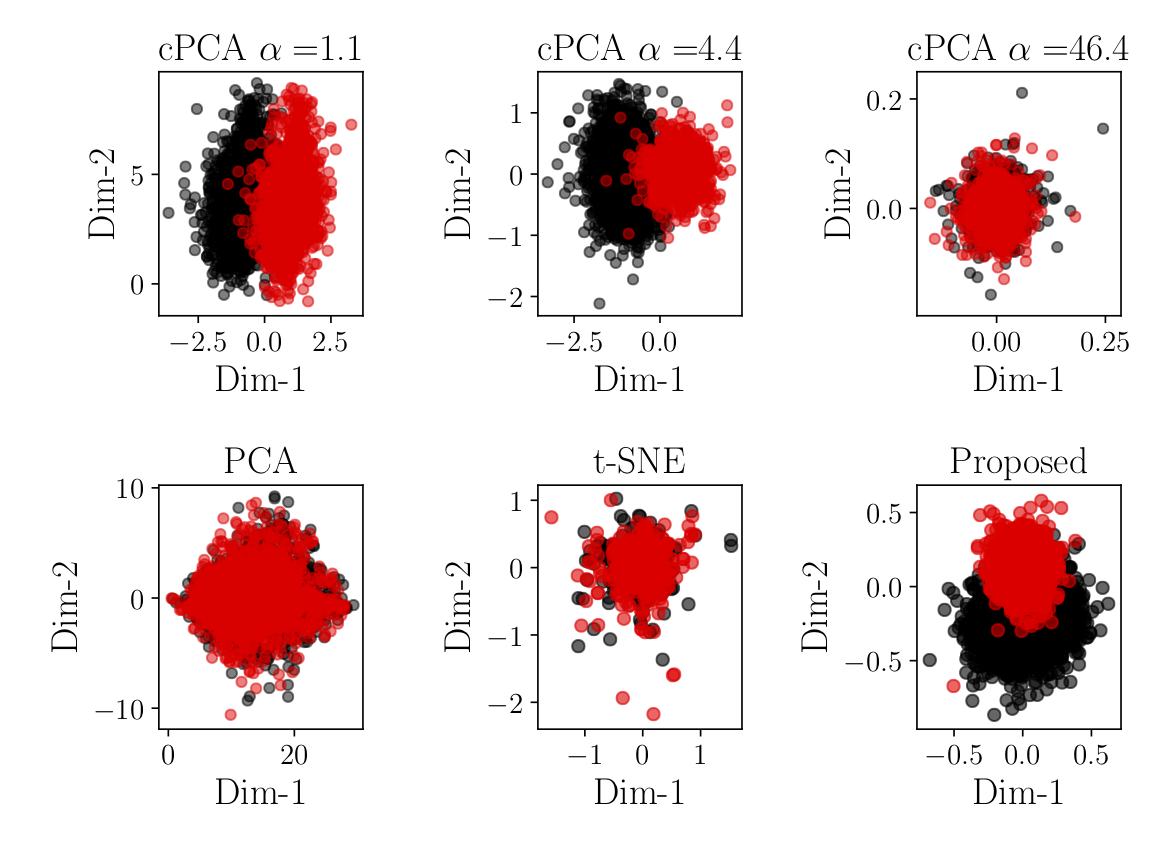}
    \caption{The performance of different dimensionality reduction techniques on the ``MNIST over Grass'' dataset illustrated in Fig.~\ref{fig:mnist_target_samples}. In all the plots, the black markers represent the digit $0$ while the red markers represent the digit $1$. The top row shows the result of executing the cPCA algorithm for different values of $\alpha$, the bottom-left plot shows the output of the traditional PCA algorithm on the target dataset, the bottom-center plot shows the output of the t-SNE algorithm, and the bottom-right plot shows the output of the cPCA++ method.}
    \label{fig:mnist}
\end{figure}

In Fig.~\ref{fig:mnist}, we evaluate the performance of the dimensionality reduction techniques on the target images illustrated in Fig.~\ref{fig:mnist_target_samples}. It can be seen that both traditional PCA and t-SNE have difficulty clustering the two digits, due to the high variance background masking the fine structure of the digits. On the other hand, the cPCA and cPCA++ methods are able to achieve some separation between the two digits, although in both cases there is some overlap. Again, note that cPCA++ does not require a hyperparameter sweep. 

\subsubsection{Mice Protein Expression}
\label{sec:mice_protein}
\begin{figure}[ht]
    \centering
    \includegraphics[width=1.02\columnwidth]{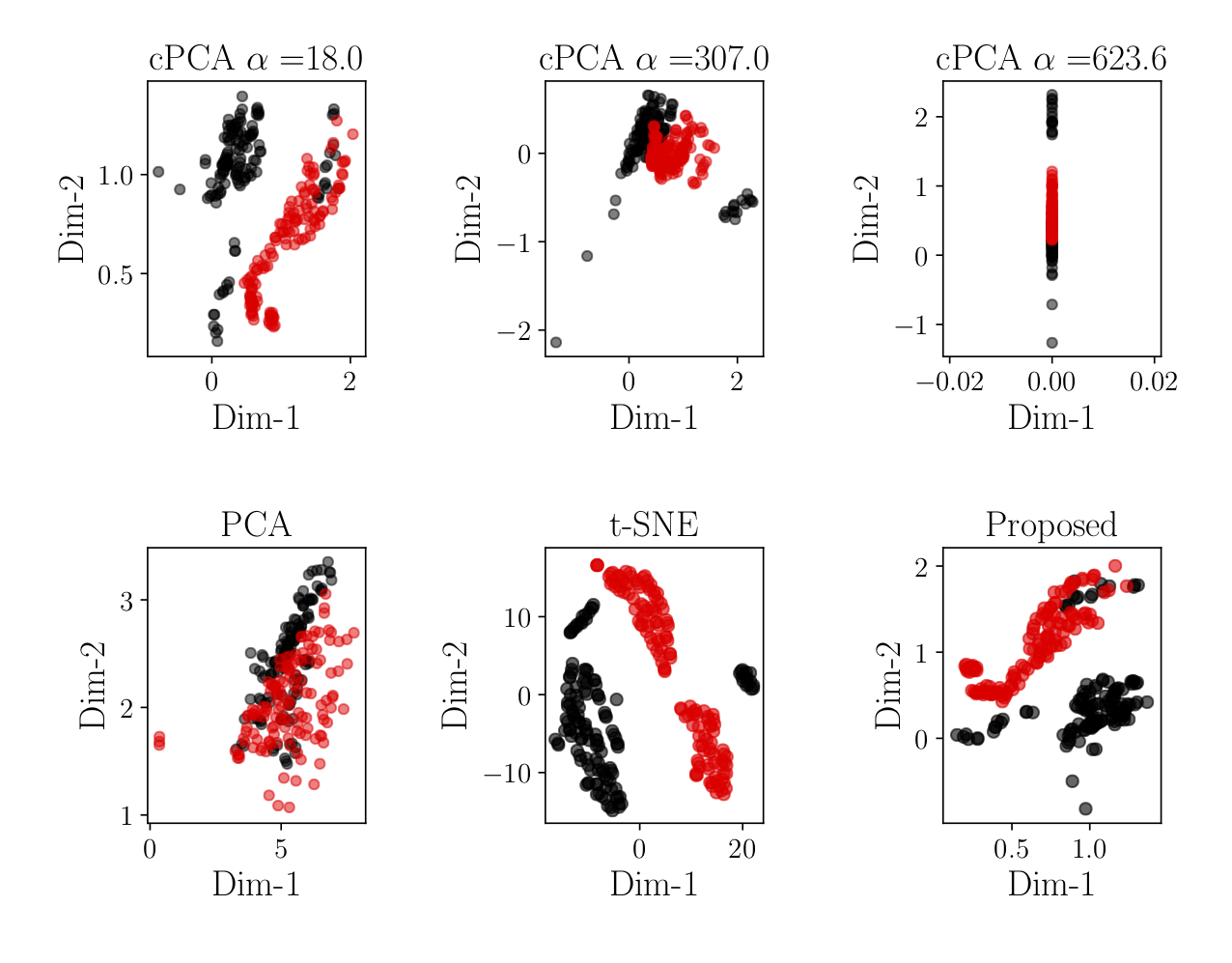}
    \caption{The performance of different dimensionality reduction techniques on the Mice Protein Expression dataset \cite{Mice}. In all the plots, the black markers represent the non-Down Syndrome mice while the red markers represent the Down Syndrome mice.}
    \label{fig:mice}
\end{figure}

This next example utilizes the public dataset from \cite{Mice}. This dataset was also used in \cite{CPCA}. In this dataset, we have access to protein expression measurements from mice that have received shock therapy and mice that have not. We would like to analyze the foreground dataset (mice with shock therapy applied) to attempt to cluster mice that have developed Down Syndrome from those that have not. We also have access to a background dataset, which consists of protein expression measurements from mice that did not experience shock therapy. In Fig.~\ref{fig:mice} we illustrate the results of the different feature reduction techniques. It can be seen that traditional PCA is unable to effectively cluster the data. On the other hand, the t-SNE, cPCA, and cPCA++ methods are all able to achieve some separation of the data. Although t-SNE was able to achieve good separation in this particular example, we re-iterate the fact that t-SNE is not an appropriate method to be used to extract \emph{filters} or function transformations from the high-dimensional space to the lower-dimensional space as the algorithm must be re-executed with new samples in order to map previously unmapped data. There is one other significant observation about this example. As indicated in the main paper, there is a very small number of samples in the background dataset. This caused the background covariance matrix, $\R_b$, to be rank-deficient. For this reason, diagonal loading was employed to ensure that the background covariance matrix is invertible.

\subsubsection{MHealth Measurements}
\begin{figure}[ht]
    \centering
    \includegraphics[width=1\columnwidth]{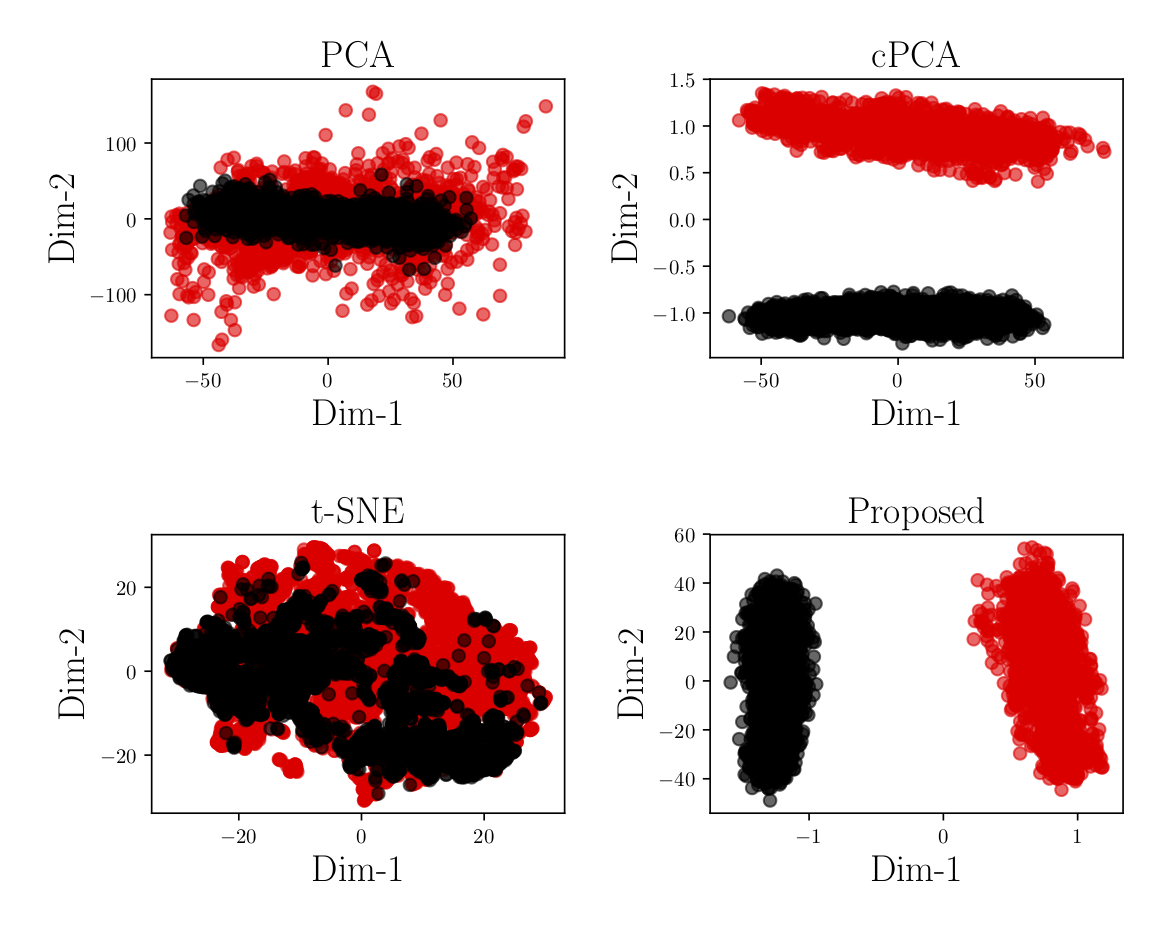}
    \caption{Clustering result of the MHealth Dataset for performing squats and cycling. In all the plots, the red markers denote squatting activity while the black markers denote cycling. In the top-left and bottom-left subplots, we see that traditional PCA and t-SNE are incapable of separating the two activities, respectively. In the top-right and bottom-right subplots, we see that cPCA and cPCA++ are capable of clustering the activities, respectively. Note that we show the optimal cPCA result (after performing the parameter sweep). }
    \label{fig:MHealth}
\end{figure}

In the next experiment (also conducted in \cite{CPCA}), we use the IMU data from \cite{IMU}. In this dataset, a variety of sensors are used to monitor subjects that are performing squats, jogging, cycling, or lying still. The sensors include gyroscopes, accelerometers, and electrocardiograms (EKG). In the top-left subplot of Fig.~\ref{fig:MHealth}, we see that performing traditional PCA on a dataset of subjects that are performing squats or cycling does not yield a reasonable clustering for those two activities. On the other hand, the optimal contrast parameter was used for cPCA, and the optimal cPCA result is illustrated in the top-right subplot of Fig.~\ref{fig:MHealth}. For both the cPCA and cPCA++ methods, the background covariance matrix $\R_b$ was obtained from subjects lying still. We observe that the cPCA algorithm was able to cluster the two activities effectively. The t-SNE output is then shown in the bottom-left subplot of the figure. We see that t-SNE was unable to distinguish the two activities. Finally, the output of the cPCA++ method is shown in the bottom-right subplot of the figure. We see that cPCA++ is able to effectively cluster the two activities, without the need for a parameter sweep. 

\subsubsection{Single Cell RNA-Seq of Leukemia Patient}
\label{sec:rna_seq}
\begin{figure}[ht]
    \centering
    \includegraphics[width=1\columnwidth]{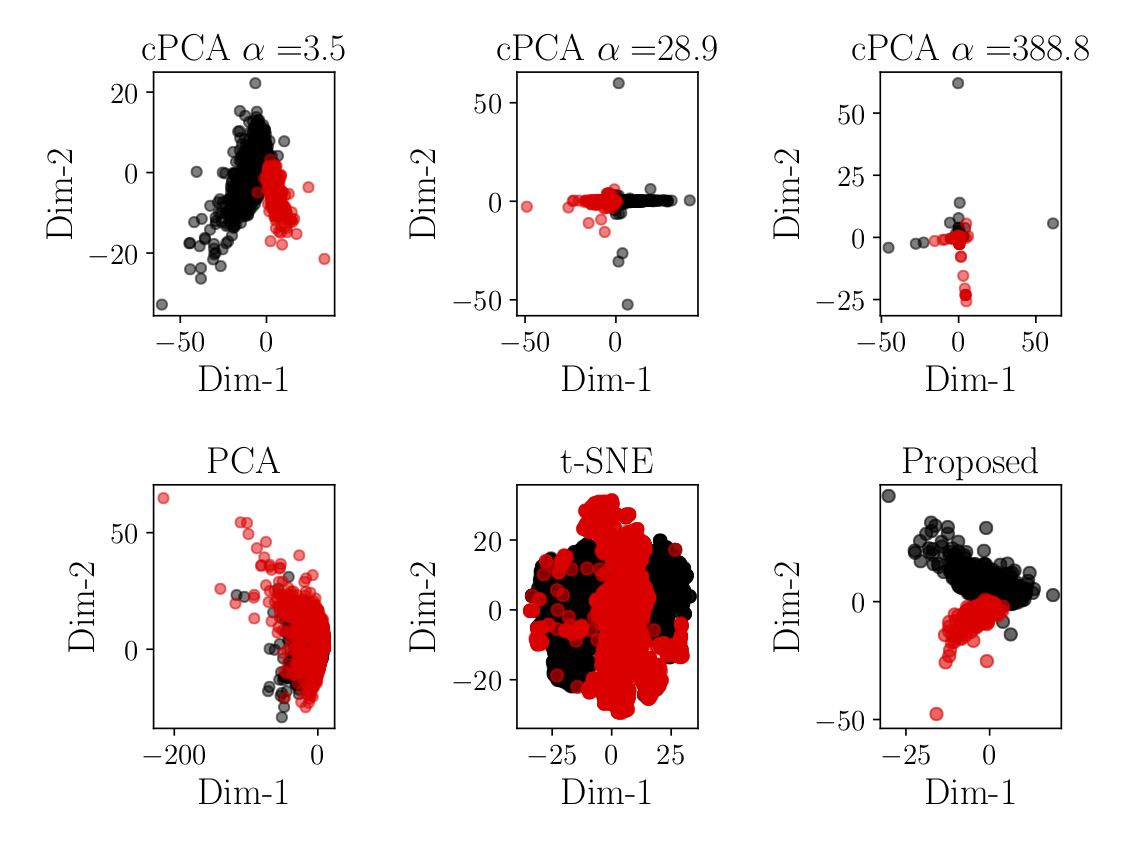}
    \caption{Dimensionality reduction result for the Single Cell RNA-Seq of Leukemia Patient example. In all plots, the black markers denote pre-transplant samples while the red markers denote post-transplant samples. The top row illustrates the output of the cPCA algorithm for varying values of $\alpha$, the bottom-left plot shows the output of traditional PCA, the bottom-center plot shows the output of the t-SNE algorithm, while the bottom-right plot shows the output of the cPCA++ method. We observe that the cPCA (for some values of $\alpha$) and cPCA++ methods yield the best clustering for this dataset.}
    \label{fig:single_cell}
\end{figure}

Our final dataset in this section was obtained from \cite{single-cell}, and this analysis was also conducted in \cite{CPCA}. The target or foreground dataset contains single-cell RNA expression levels of a mixture of bone marrow mononuclear cells (BMMCs) from a leukemia patient before and after stem cell transplant, and the goal is to separate the pre-transplant samples from the post-transplant samples. The original features were reduced to only 500 genes based on the value of the variance divided by the mean. Once these features are obtained, we execute the cPCA algorithm for different values of $\alpha$ and plot the result in the top row of Fig.~\ref{fig:single_cell}. In this example, the background covariance matrix $\R_b$ was obtained from RNA-Seq measurements from a healthy individual's BMMCs, for both the cPCA and cPCA++ methods. As noted in \cite{CPCA}, this should allow the dimensionality reduction methods to focus on the novel features of the target dataset as opposed to being overwhelmed by the variance due to the heterogeneous population of cells as well as variations in experimental conditions. We observe that the data is close to being separable for the contrast value of $\alpha=3.5$. Next, we plot the traditional PCA output in the bottom-left subplot of Fig.~\ref{fig:single_cell}, and observe that it is incapable of clustering the data. Next, in the bottom-center plot, we illustrate the output of the t-SNE algorithm, and see that while it does achieve some separation of the data, it is not on par with the cPCA algorithm. Finally, the cPCA++ output is illustrated in the bottom-right subplot of Fig.~\ref{fig:single_cell}, and we see that it is able to obtain a similar clustering performance as compared to the optimal cPCA clustering result. Note that cPCA++ does not require a parameter sweep. The covariance matrix $\R_b$ was rank deficient in this example, so we utilized diagonal loading in order to make it invertible.

\subsection{Computational Time Performance Comparison}
\label{sec:timing}
In this section, we will examine the computational time requirements for the various datasets studied in Sec.~\ref{sec:comparison_to_cPCA}. We neglect measuring the computation time required by the traditional PCA algorithm since it will be very similar to the cPCA++ algorithm and since traditional PCA is not an effective algorithm for discriminative dimensionality reduction, as we saw in the results of Sec.~\ref{sec:comparison_to_cPCA}. As shown in Table \ref{tbl:datasets_algorithm_performance}, the proposed cPCA++ method is significantly faster than both the cPCA method and the t-SNE method. The reason why the cPCA++ method is on average (i.e., averaged over the experiments outlined in Sec.~\ref{sec:comparison_to_cPCA}) $51$ times faster than the cPCA method is that while both methods require the eigen-decomposition of similarly sized covariance matrices, the cPCA algorithm requires a \emph{sweep} over the contrast hyper-parameter $\alpha$ (i.e., multiple eigen-decompositions). On the other hand, the cPCA++ method does not require this hyper-parameter tuning since the scale discrepancy is automatically resolved. It is also notable to see that cPCA++ is on average $428,654$ times faster than t-SNE. The t-SNE method has a very high running time due to the fact that it works by finding relationships between all the points in a dataset, and not through the determination of compact parameters. In addition, it is not very well suited for dimensionality reduction when future data samples will be obtained (since it has to be re-executed on the entire new dataset). 

\begin{table}[!t]
\renewcommand{\arraystretch}{1.7}
\caption{Time required for the different algorithms to perform the required dimensionality reduction for the various datasets studied in Sec.~\ref{sec:comparison_to_cPCA}. All times listed in the table are in seconds. \textbf{Boldface} is used to indicate shortest runtimes and average cPCA++ speedup.}
\label{tbl:datasets_algorithm_performance}
\centering
\begin{tabular}{c||c|c|c}
\hline \hline
\rowcolor[gray]{0.9} \rule[-1ex]{0pt}{4ex} \textbf{Example} & cPCA & t-SNE & cPCA++\\ 
\hline
\rule[-1ex]{0pt}{4ex} \textbf{Synthetic} & $0.062$ & $6.62$ & $\mathbf{0.0017}$  \\ 
\hline 
\rule[-1ex]{0pt}{4ex} \textbf{MNIST over Grass} & $26$ & $811$ & $\mathbf{1.00}$  \\ 
\hline 
\rule[-1ex]{0pt}{4ex} \textbf{Mice Protein Expression} & $0.13$ & $2.23$ & $\mathbf{0.0033}$  \\ \hline
\rule[-1ex]{0pt}{4ex} \textbf{MHealth Measurements} & $0.091$ & $1388$ & $\mathbf{0.00065}$  \\
\hline
\rule[-1ex]{0pt}{4ex} \textbf{RNA-Seq  of  Leukemia  Patient} & $11.70$ & $2155$ & $\mathbf{0.86}$  \\
\hline \hline
\rule[-1ex]{0pt}{4ex} \textbf{Average cPCA++ Speedup} & $\mathbf{51}$\textbf{x} & $\mathbf{428654}$\textbf{x} & $\mathbf{1}$\textbf{x}  \\
\hline \hline
\end{tabular} 
\end{table}

\section{The cPCA++ Framework for Image Splicing Localization}
\label{sec:image_splice_localization_setup}
In this section, we describe the general framework of the cPCA++ approach for image splicing localization. 
In this approach, we focus on detecting the spliced boundary or edge.
\begin{figure}
\centering
\begin{subfigure}{.4\columnwidth}
  \centering
  \includegraphics[width=1\columnwidth]{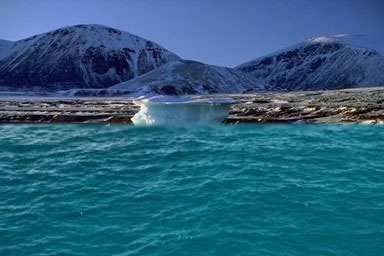}
  \caption{Probe Image}
  \label{ambiguity_probe}
\end{subfigure}\quad%
\begin{subfigure}{.4\columnwidth}
  \centering
  \includegraphics[width=1\columnwidth]{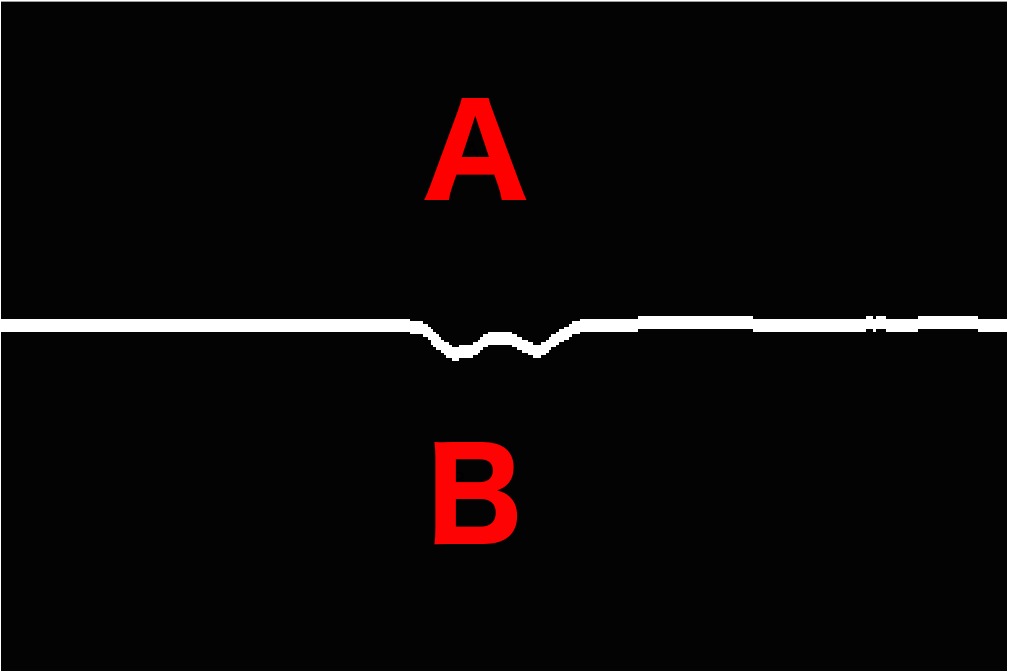}
  \caption{Spliced Edge}
  \label{ambiguity_edge}
\end{subfigure}
\caption{An example illustrating that there is an ambiguity in how one labels the spliced and authentic regions in a given probe image. The image on the right is a ground truth mask highlighting the spliced edge or boundary. The color white is used to denote a pixel belonging to the spliced boundary, while the color black is used to denote a pixel not belonging to the spliced boundary. It is possible to label region A as the spliced region and region B as the authentic region, or vice versa. }
\label{ambiguity}
\end{figure}
We take this approach of determining the spliced boundary rather than the spliced region or surface because there is an ambiguity in the determination of which image is the donor and the host. For example, consider Figure \ref{ambiguity}, which shows a probe image and the corresponding spliced boundary ground truth mask. Please note that we use the color white to denote a pixel belonging to the spliced boundary, and the color black to denote a pixel not belonging to the spliced boundary. It is possible to consider region A as the spliced region and region B as the authentic region, or vice versa. Thus, there is an ambiguity in how one labels the spliced and authentic regions in a given manipulated image. However, this ambiguity is resolved if we instead consider the spliced boundary (rather than the spliced surface/region).
  
The ``training'' phase consists of two main steps: 1) collect labeled or ``training'' samples/patches, 2) build a feature extractor based on the training samples. As we will see later, the cPCA++ approach does not require the training of a classifier, such as a support vector machine or random forest. However, we will still refer to patches used in this phase as ``training'' samples, since we are using their labeled information (i.e., spliced vs. authentic edges). In the testing phase, we are given a new image (i.e., not seen during the training phase), and the goal is to output a probability map that indicates the probability that each pixel belongs to the spliced boundary. The testing phase consists of the following main steps: 1) divide a given test image into overlapping patches, 2) extract the feature vector for each test patch, 3) obtain the probability output for each test patch, and 4) reconstruct the final output map for the given test image. The training and testing phases are elaborated below.

In the training phase, the first step is to collect labeled or ``training" samples/patches from a given set of images. In order to do this, we need access to two masks for each training image: 1) the spliced surface ground truth mask and 2) a mask which highlights the superset of all edges/boundaries in the image
(i.e., spliced boundaries as well as natural/authentic boundaries), referred to as the edge detection output mask. Note that although we utilize the surface ground truth mask in selecting the training patches, the output of the cPCA++ method will be edge-based (i.e., the output will represent an estimate of the true spliced edge/boundary).  This will be explained in more detail in the next paragraph. We utilized the CASIA v2.0 dataset for training purposes. For the CASIA v2.0 dataset, the surface ground truth masks are not provided. We generated the ground truth masks using the provided reference information. In particular, for a given spliced image in the CASIA dataset, the corresponding donor and host images are provided, and we used this information to generate the ground truth mask. The second mask that is needed for collecting the training patches is the edge detection output mask. We utilized structured edge detection for detecting spliced and natural edges \cite{Dollar:2013}.

\begin{figure}
\centering
\includegraphics[width=1\columnwidth]{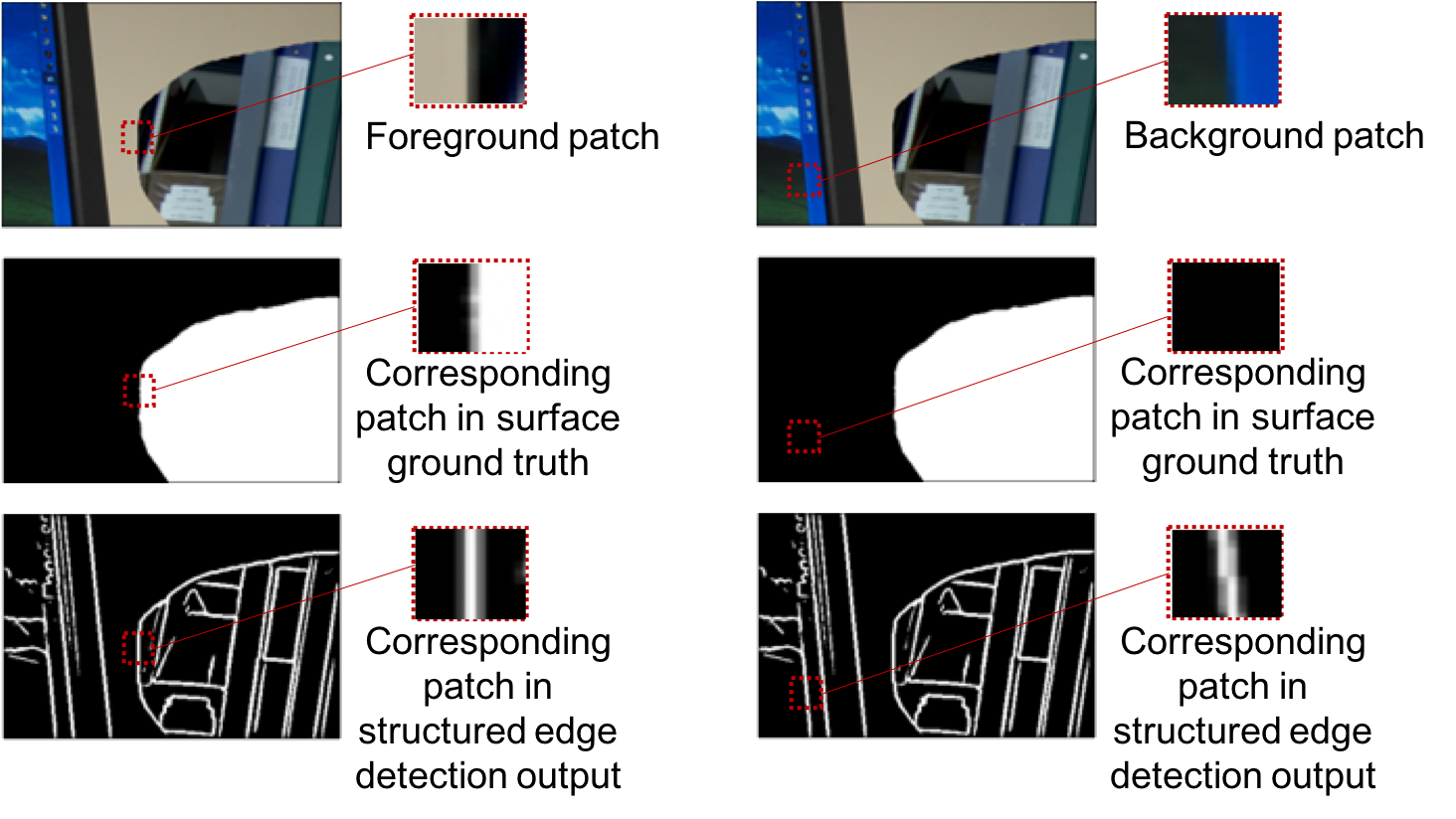}
\caption{Example of foreground and background patches.}
\label{fig:both_patches}
\end{figure}

Once we have these two masks, we can proceed to collect the training patches. Each training image is divided into overlapping patches, and we then select target/foreground and background samples in the following manner. We would like target/foreground samples to be those that contain a boundary between a spliced region and an authentic region (referred to as a spliced boundary). In order to achieve this, we utilize the surface ground truth masks, and we select foreground samples to be those that have a splicing fraction that lies in a certain range (e.g., 30-70 percent of the patch is a spliced area/surface). On the other hand, background samples do not contain a boundary between a spliced and authentic region. We impose an additional constraint on the background samples so that they contain a minimum amount of authentic edges (as specified by the structured edge detector). Figure \ref{fig:both_patches} shows an example of a foreground patch and a background patch. Note that although we are using the surface ground truth masks in selecting foreground and background samples, what is actually important is the spliced boundary/edge (or lack thereof).

The next step is to build a feature extractor based on the covariance matrices of the foreground and background training samples. Suppose we have collected $N_b$ background patches and $N_f$ foreground patches, and let the size of each patch be $n\times n\times c$, where $c$ is the number of channels. Since we are working with RGB images, $c$ will be equal to $3$. Each training patch can be flattened so that it is a $M\times 1$ vector, where $M$ is equal to $cn^{2}$ (or 3$n^{2}$ if $c = 3$). We can represent the flattened background and foreground patches as data matrices $\widetilde{\Z}_b\in\mathbb{R}^{M\times N_b}$ and $\widetilde{\Z}_f\in\mathbb{R}^{M\times N_f}$, respectively, such that different columns of $\widetilde{\Z}_b$ (or $\widetilde{\Z}_f$) correspond to different training samples. 

To see why it is necessary to use the cPCA++ method as opposed to traditional PCA for the image splicing localization task, consider the top 50 eigenvectors of the covariance matrix of $\widetilde{\Z}_b$ (i.e.,  authentic edges or background) and the top 50 eigenvectors of the covariance matrix of $\widetilde{\Z}_f$ (i.e., spliced edges or target/foreground). We compute the power of the foreground eigenvectors ($\U_f$) after projecting them onto the background subspace (spanned by $\U_b$) to quantify the similarity in the subspaces spanned by $\widetilde{\Z}_b$ and $\widetilde{\Z}_f$:
$\|\P_{\U_b} \U_f\|_F^2/\| \U_f\|_F^2 = 93.25\%$. What this indicates is that the subspaces spanned by the top $50$ principal components of each dataset ($\U_f$ and $\U_b$) are mostly overlapping---hinting at the fact that it is likely that any classifier will be overwhelmed by the similarity of the features of these two datasets. For this reason, we will utilize the cPCA++ method instead to alleviate this situation and perform \emph{discriminative} feature extraction.

We thus utilize Alg.~\ref{alg:cPCA++}, described in Sec.~\ref{sec:cPCA++}, to build the feature extractor. We first center the data matrices $\widetilde{\Z}_b$ and $\widetilde{\Z}_f$ by subtracting their respective means. We then calculate the second order statistics $\R_b = \frac{1}{N_b} \Z_b \Z_b^\T$ and $\R_f = \frac{1}{N_f} \Z_f \Z_f^\T$, and then perform eigenvalue decomposition on $\Q = \R_b^{-1} \R_f$. If the matrix $\R_b$ is rank-deficient, we utilize diagonal loading in order to make it invertible. We then compute the top $K$ right-eigenvectors $\F$ of $\Q$. The matrix $\F \in\mathbb{R}^{M\times K}$ is used to extract features during the testing phase, and is referred to as the transform matrix.

In the testing phase, we divide each test image into overlapping patches. Similar to what was done during the training phase, we can flatten the test patches and represent them as a matrix $\widetilde{\Z}_t\in\mathbb{R}^{M\times N_{t}}$. We then utilize Algorithm \ref{alg:classify_patch} to obtain a probability output for each patch. The motivation behind Algorithm  \ref{alg:classify_patch} is explained as follows. The cPCA++ method will naturally find filters that will yield a small norm for the reduced dimensionality features of background samples and a larger norm for foreground samples. This means that once the filters are obtained, they are expected to yield small valued output when applied to background patches and larger valued output when applied to foreground patches. This presents us with a very simple and efficient technique for obtaining the output: simply measure the output power after dimensionality reduction and then convert the raw value to a probability-based one. The full output map for a given test image can then be reconstructed by averaging the contributions from overlapping regions. After reconstruction, we perform an element-wise multiplication of the output map and the structural edge detection output mask. The reason for doing this is explained as follows. In this task, we are attempting to classify the spliced edge. Therefore, if the structural edge detector labeled a given pixel as non-edge (i.e., neither an authentic edge nor a spliced edge), then the corresponding pixel value in the cPCA++ output map should automatically be set to zero.

\begin{algorithm}
   \caption{Algorithm for obtaining an output for each test patch}
   \label{alg:classify_patch}
\begin{algorithmic}
   \STATE {\bfseries Inputs:} Test data matrix $\widetilde{\Z}_t\in\mathbb{R}^{M\times N_{t}}$; transform matrix $\F \in\mathbb{R}^{M\times K}$, obtained during the training phase
   \vspace{2mm}
   \begin{enumerate}
       \STATE Center the matrix $\widetilde{\Z}_t$ to obtain $\Z_t$
       \STATE Compute $\Y_t = \F^\T \Z_t$, where each column of $\Y_t$ contains the reduced-dimension feature vector for a given test patch
       \STATE Compute the vector $\v \in\mathbb{R}^{N_t\times 1}$ where each element $\v_i \triangleq \|\Y_{t,i}\|_2^2$ for $1\leq i\leq N_t$ is the squared-$L_{2}$-norm of a given column of $\Y_t$
       \STATE Convert $\v$ to a probability vector $\w$ by computing $\w = \frac{\v}{\max(\v)}$
   \end{enumerate}
   \STATE{\bfseries Return:} the vector $\w \in\mathbb{R}^{N_t\times 1}$
\end{algorithmic}
\end{algorithm}

\section{Experimental Analysis}
\label{sec:experimental-analysis}

\subsection{Evaluation/Scoring Procedure}
\label{subsec:scoring-procedure}
In this subsection, we explain the procedure used to score the output maps from the cPCA++ method, as well as from the Multi-task Fully Convolutional Network (MFCN)-based method \cite{salloum2018image} we compared against. The MFCN-based method is a state-of-the-art approach that also outputs an edge-based probability map, and thus can be directly compared with the proposed cPCA++ method. The MFCN achieved the highest score in the splicing localization task in the 2017 Nimble Challenge, and the second highest score in the 2018 Media Forensics Challenge (which are part of the DARPA MediFor program\footnote{https://www.darpa.mil/program/media-forensics}). We evaluated the performance of the methods using the $F_{1}$ and Matthews Correlation Coefficient ($MCC$) metrics, which are popular per-pixel localization metrics. As noted previously, the edge-based approach that we propose avoids the ambiguity in the labeling of the spliced and authentic surfaces/regions. Because of the ambiguity in the surface-based labeling, most surface-based scoring procedures score the original output map as well as the inverted version of the output map, and select the one that yields the best score. However, in edge-based scoring procedures, there is no need to invert the edge-based output map, and the output can be scored directly. 
For a given spliced image, the $F_{1}$ metric is defined as
\[
F_{1}=\frac{2TP}{2TP+FN+FP},
\]
where $TP$ represents the number of pixels classified as true positive, $FN$ represents the number of pixels classified as false negative, and $FP$ represents the number of pixels classified as false positive. In this case, the target/foreground patches can be viewed as the positive samples and the background patches can be viewed as the negative samples. The $F_{1}$ metric ranges in value from $0$ to $1$, with a value of $1$ being the best. For a given spliced image, the $MCC$ metric is defined as
\textcolor{black}{
\[
MCC\!=\!\frac{TP\cdot TN-FP\cdot FN}{\sqrt{(TP\!+\!FP)(TP\!+\!FN)(TN\!+\!FP)(TN\!+\!FN)}}.
\]}The $MCC$ metric ranges in value from $-1$ to $1$, with a value of $1$ being the best.

\subsection{Experimental Results}
\label{subsec:experimental-results}
We first compared the training times of the proposed cPCA++ method and the MFCN-based method, and found the proposed method to be significantly more efficient. The MFCN-based method requires a training time of approximately 11.5 hours on a NVIDIA GeForce GTX Titan X GPU, while the proposed cPCA++ method only requires a training time of approximately 1.5 hours on CPU (Intel Xeon Gold 6126). Please note that the cPCA++ code has not yet been optimized to make full utilization of the multi-core processor. Next, we evaluated the proposed cPCA++ method on the Columbia and Nimble WEB datasets and compared its performance with the MFCN-based method. Tables \ref{mcc_scores} and \ref{f1_scores} show the Matthews Correlation Coefficient ($MCC$) and $F_{1}$ scores, respectively. It can be seen that the cPCA++ method yields higher scores (in terms of both $MCC$ and $F_{1}$), as compared to the MFCN-based method. 

\begin{figure}
\centering
\includegraphics[width=1\columnwidth]{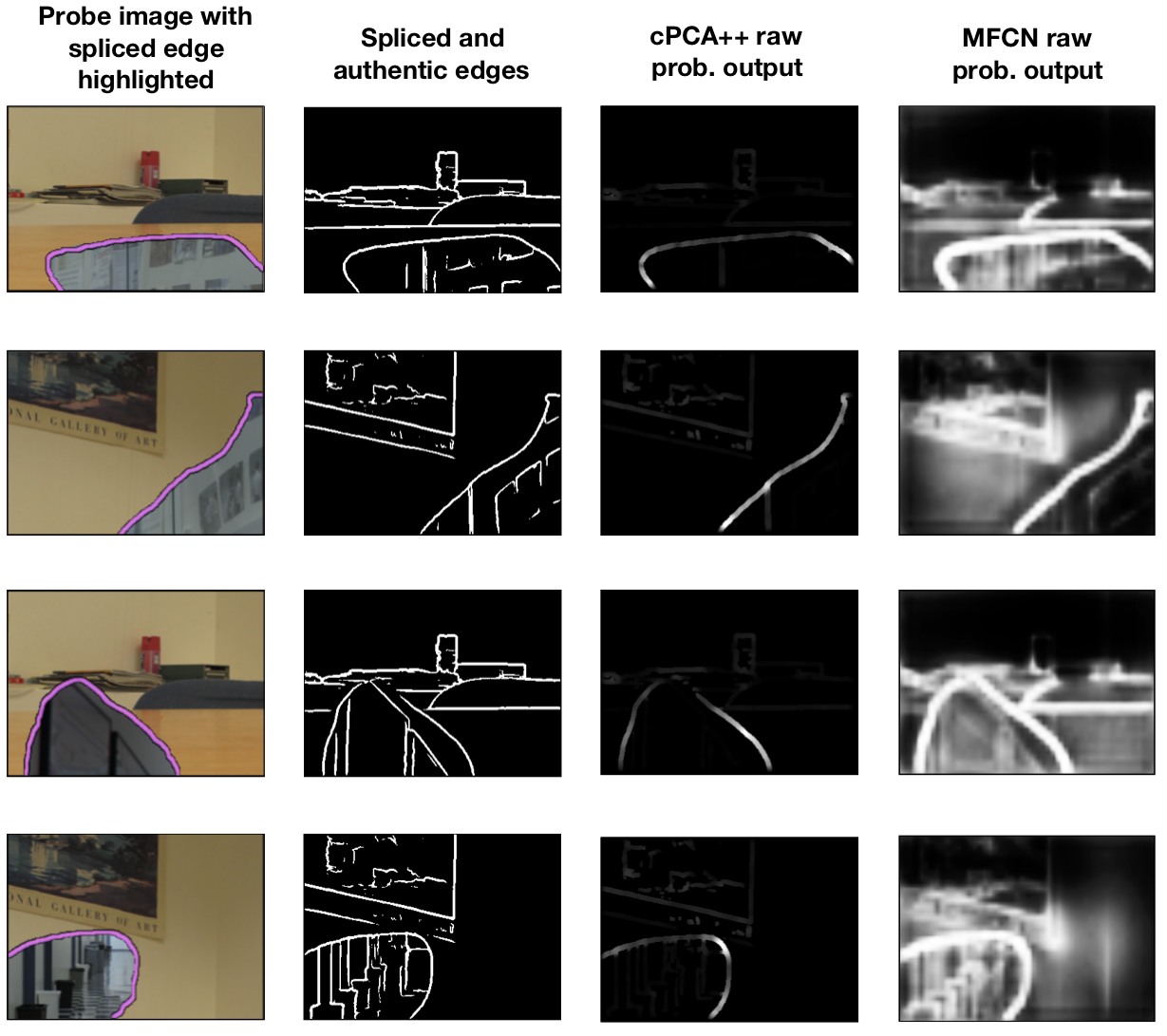}
\caption{Localization Output Examples from Columbia Dataset. Each row shows (from left to right): the manipulated/probe image with the spliced edge highlighted in pink, the structural edge detection output mask highlighting both spliced and authentic edges,  the cPCA++ raw probability output, and the MFCN-based raw probability output.}
\label{fig:columbia_examples}
\end{figure}

\begin{figure}
\centering
\includegraphics[width=1\columnwidth]{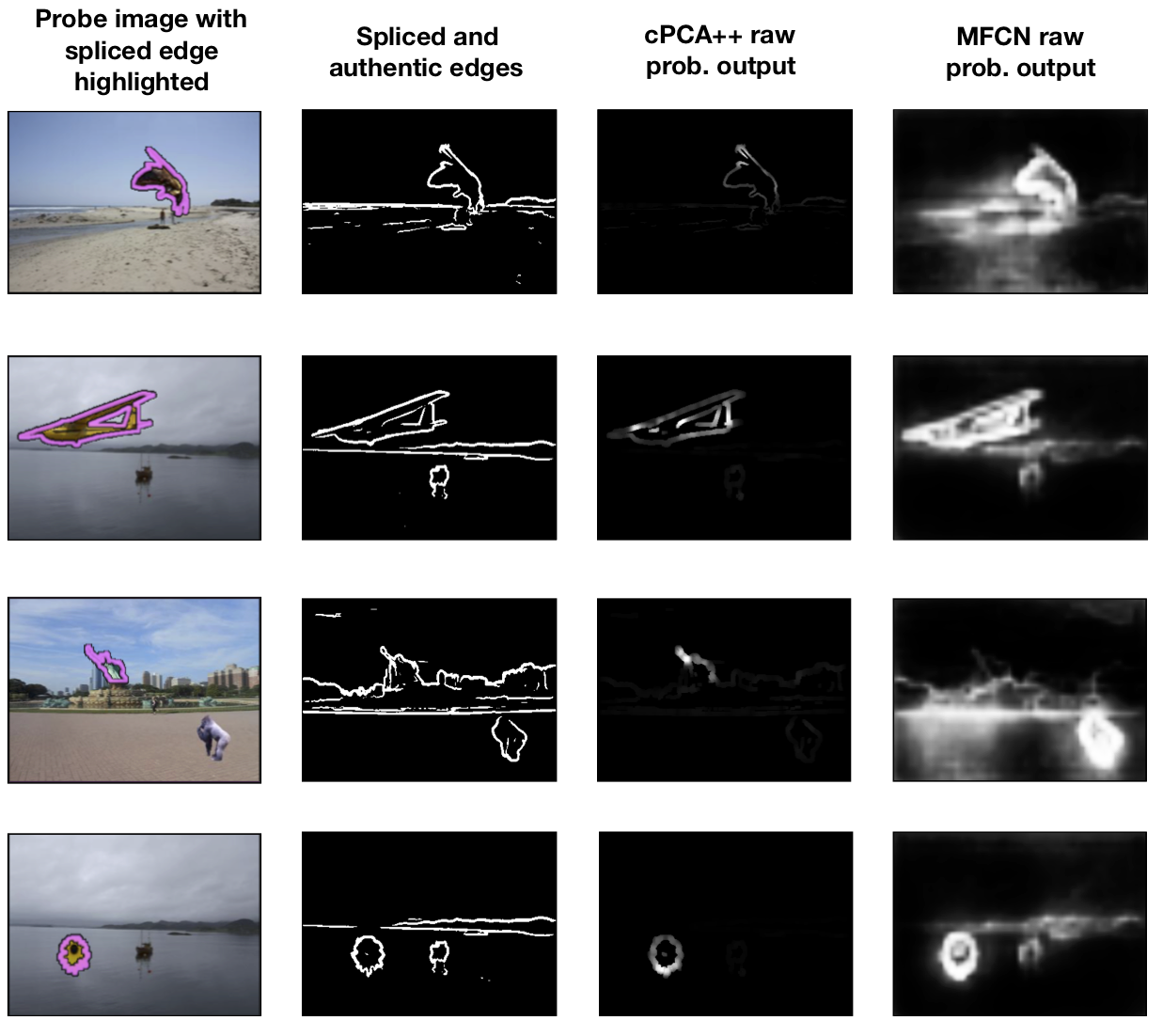}
\caption{Localization Output Examples from Nimble WEB Dataset. Each row shows (from left to right): the manipulated/probe image with the spliced edge highlighted in pink, the structural edge detection output mask highlighting both spliced and authentic edges,  the cPCA++ raw probability output, and the MFCN-based raw probability output.}
\label{fig:nc2016web_examples}
\end{figure}

Figures \ref{fig:columbia_examples} and \ref{fig:nc2016web_examples} show multiple examples of localization output from the Columbia and Nimble WEB datasets, respectively. Each row shows (from left to right) a manipulated or probe image with the spliced edge highlighted in pink, the structural edge detection output mask highlighting both spliced and authentic edges, the cPCA++ raw probability output map, and the MFCN-based raw probability output map. In these figures, it can be seen that the cPCA++ method yields a finer localization output than the MFCN-based method.

\begin{table}[!t]
\renewcommand{\arraystretch}{1.7}
\caption{Edge-based $MCC$ Scores on Columbia and Nimble WEB Datasets. \textbf{Boldface} is used to emphasize best performance.}
\label{mcc_scores}
\centering
\begin{tabular}{c||c|c}
\hline \hline
\rowcolor[gray]{0.9} \rule[-1ex]{0pt}{4ex} \textbf{Dataset} & cPCA++ & MFCN\\ 
\hline
\rule[-1ex]{0pt}{4ex} \textbf{Columbia} & $\mathbf{0.385}$ & $0.329$  \\ 
\hline 
\rule[-1ex]{0pt}{4ex} \textbf{Nimble WEB} &  $\mathbf{0.388}$ & $0.297$  \\ 
\hline \hline
\end{tabular} 
\end{table}

\begin{table}[!t]
\renewcommand{\arraystretch}{1.7}
\caption{Edge-based $F_{1}$ Scores on Columbia and Nimble WEB Datasets. \textbf{Boldface} is used to emphasize best performance.}
\label{f1_scores}
\centering
\begin{tabular}{c||c|c}
\hline \hline
\rowcolor[gray]{0.9} \rule[-1ex]{0pt}{4ex} \textbf{Dataset} & cPCA++ & MFCN\\ 
\hline
\rule[-1ex]{0pt}{4ex} \textbf{Columbia} & $\mathbf{0.359}$ & $0.312$  \\ 
\hline 
\rule[-1ex]{0pt}{4ex} \textbf{Nimble WEB} &  $\mathbf{0.376}$ & $0.273$  \\ 
\hline \hline
\end{tabular} 
\end{table}

\section{Conclusion}
\label{sec:conclusion}
In conclusion, we proposed cPCA++, which is a new technique for discovering discriminative features in high-dimensional data. The proposed approach is able to discover structures that are unique to a target dataset, while at the same time suppressing uninteresting high-variance structures present in both the target dataset and a background dataset. The proposed cPCA++ approach is compared with a recently proposed algorithm, called contrastive PCA (cPCA), and we show that cPCA++ achieves similar discriminative performance in a wide variety of settings, even though it eliminates the need for the hyperparameter sweep required by cPCA. Following  this discussion, the cPCA++ approach was applied to the problem of image splicing localization, in which the two classes of interest (i.e., spliced and authentic edges) are extremely similar in nature. In the context of this problem, the target dataset contains the spliced edges and the background dataset contains the authentic edges. We show that the cPCA++ approach is able to effectively discriminate between the spliced and authentic edges. The resulting method was evaluated on the Columbia and Nimble WEB splicing datasets. The proposed method achieves scores comparable to the state-of-the-art Multi-task Fully  Convolutional Network  (MFCN),  and  it  does  so very  efficiently–--without  the  need to iteratively update filter weights via stochastic gradient descent and backpropagation, and without the need to train a classifier.

\appendices
\section{Derivation of Filters for Synthetic Example}
\label{app:synthetic}
The sample covariance of the background dataset specified in \eqref{eq:synthetic_bg} will be:
\begin{align}
    \R_b &\approx \left[\begin{array}{ccc}
    3\boldsymbol{I}_{10\times 10} & \boldsymbol{0}_{10\times 10} & \boldsymbol{0}_{10\times 10}\\
    \boldsymbol{0}_{10\times 10} & \boldsymbol{I}_{10\times 10} & \boldsymbol{0}_{10\times 10} \\
    \boldsymbol{0}_{10\times 10} &  \boldsymbol{0}_{10\times 10} & 10\boldsymbol{I}_{10\times 10}
    \end{array}\right] \\
    &=  \left[\begin{array}{ccc}
    3 & 0 & 0\\
    0 & 1 & 0 \\
    0 & 0 & 10
    \end{array}\right] \otimes \boldsymbol{I}_{10\times 10}
\end{align}
where the $\A\otimes \B$ operator denotes the Kronecker product of the matrices $\A$ and $\B$.
On the other hand, the sample covariance of the target/foreground dataset is given by (when the classes are equally likely):
\begin{align}
    \R_f &\approx  \left[\begin{array}{ccc}
    9  & 0 & 0\\
    0 & 2.25 & 0 \\
    0 & 0 & 0
    \end{array}\right] \otimes \mathds{1}_{10}\mathds{1}_{10}^\T + \nonumber\\
    &\quad\ \left[\begin{array}{ccc}
    1  & 0 & 0\\
    0 & 1 & 0 \\
    0 & 0 & 10
    \end{array}\right] \otimes \I_{10\times 10}.
\end{align}
Then, the matrix $\Q$ for the cPCA++ algorithm becomes:
\begin{align}
    \Q &\triangleq \R_b^{-1} \R_f \nonumber\\
        &= \left[\begin{array}{ccc}
    3 & 0 & 0\\
    0 & 2.25 & 0 \\
    0 & 0 & 0
    \end{array}\right] \otimes \mathds{1}_{10}\mathds{1}_{10}^\T  + \nonumber\\
    &\quad\ \left[\begin{array}{ccc}
    \frac{1}{3} & 0 & 0\\
    0 & 1 & 0 \\
    0 & 0 & 1
    \end{array}\right] \otimes \boldsymbol{I}_{10\times 10}. \label{eq:analysis_approximation}
\end{align}
The matrix $\Q$ turns out to be symmetric (and thus is diagonalizable) and has the following block-diagonal structure:
\begin{align}
    \Q = \mathrm{blockdiag}(\A,\B,\C)
\end{align}
where
\begin{align}
    \A &\triangleq 3 \mathds{1}_{10}\mathds{1}_{10}^\T + \frac{1}{3} \I_{10\times 10}\\
    \B &\triangleq 2.25 \mathds{1}_{10}\mathds{1}_{10}^\T + \I_{10\times 10}\\
    \C &\triangleq \I_{10\times 10}.
\end{align}
The eigenvalues of the block diagonal matrix $\Q$ are given by the eigenvalues of its diagonal blocks\footnote{This is because an eigenvalue $\lambda$ of $\Q$ by definition satisfies $|\Q-\lambda \I_{30\times 30}|=0$. However, since $\Q$ is block diagonal, we have that $|\Q-\lambda \I_{30\times 30}| = |\A-\lambda \I_{10\times 10}|\cdot |\B-\lambda \I_{10\times 10}|\cdot |\C-\lambda \I_{10\times 10}|$ \cite[p.~5]{Laub}. Clearly, if $\lambda$ is an eigenvalue of $\A$, $\B$, or $\C$, it is also an eigenvalue of $\Q$ since it will also satisfy $|\Q-\lambda \I_{30\times 30}|=0$. There are a total of $30$ such values ($10$ per block), some with multiplicity greater than one.}, i.e., if the eigen-decomposition of $\Q$ is given by $\Q = \U \D \U^\T$, then
\begin{align}
     \D &\triangleq \mathrm{blockdiag}(\D_A, \D_B, \D_C)\\
        &\stackrel{(a)}{=}  \mathrm{diag}\left(30.33,23.50,\mathds{1}_{19}^\T,\frac{1}{3}\mathds{1}_{9}^\T\right)
\end{align}
where the matrices $\D_A$, $\D_B$, and $\D_C$ contain the eigenvalues of the matrices $\A$, $\B$, and $\C$, respectively, along their diagonal and the eigenvalues are sorted in decreasing order in step $(a)$. It is then easy to verify that:
\begin{align}
    \Q \u_1 &= 30.33 \u_1\\
    \Q \u_2 &= 23.50 \u_2
\end{align}
where
\begin{align}
    \u_1 = \left[
    \begin{array}{c}
    \frac{1}{\sqrt{10}} \mathds{1}_{10}\\
     \mathbb{0}_{10} \\
     \mathbb{0}_{10}
    \end{array}
    \right], \quad\quad \u_2 = \left[
    \begin{array}{c}
    \mathbb{0}_{10}\\
    \frac{1}{\sqrt{10}} \mathds{1}_{10}\\
     \mathbb{0}_{10}
    \end{array}
    \right]
\end{align}
indicating that $\u_1$ and $\u_2$ are the principal eigenvectors of $\Q$, yielding \eqref{eq:synthetic_filters}.

\section{The cPCA++ Approach For Matrix Factorization and Image Denoising}
\label{app:denoising}
\begin{figure}[ht]
    \centering
    \includegraphics[width=1\columnwidth]{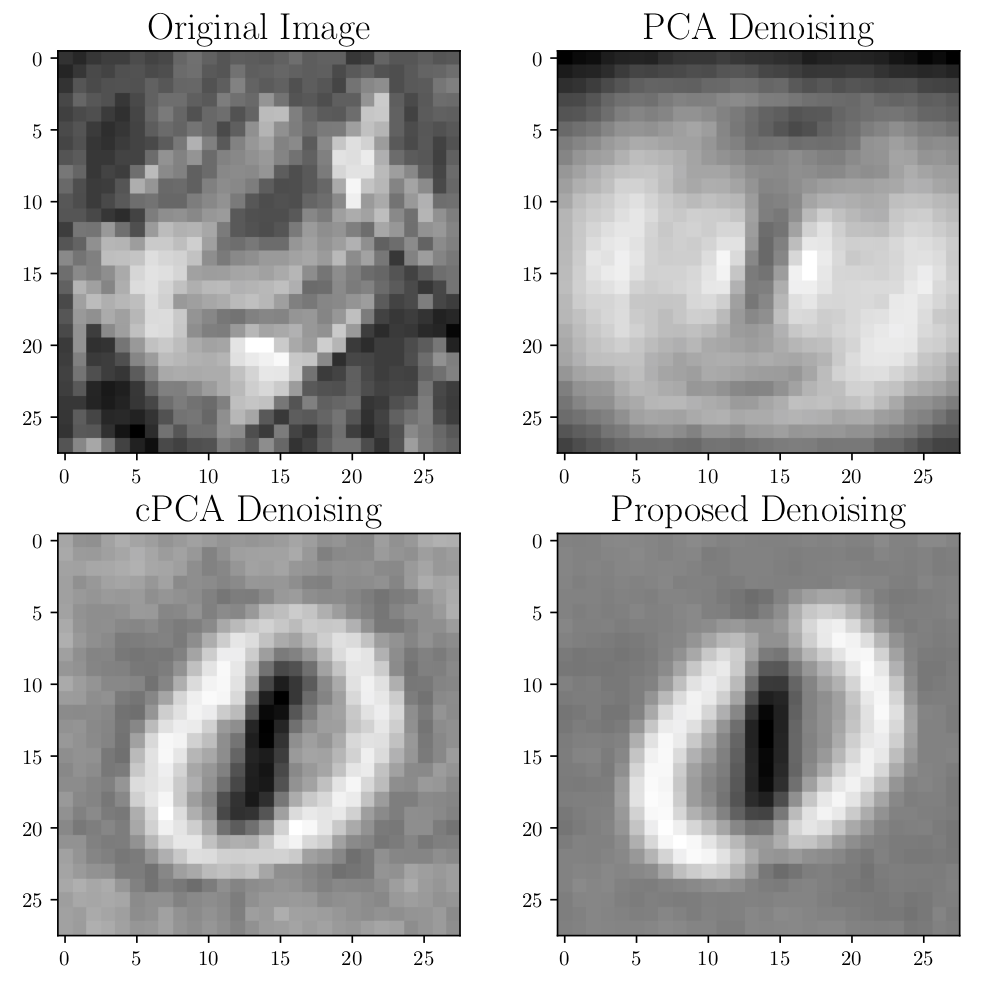}
    \caption{The denoising of the digit $0$ over grass background. Please note that we use the color white to denote pixels corresponding to the digit. On the top-left plot, we show the original noisy digit. On the top-right plot, we show the denoising achieved by traditional PCA. On the bottom-left plot, we show the denoising achieved by the cPCA algorithm. Finally, on the bottom-right plot, we show the denoising performance of the cPCA++ method, which is the low-rank approximation $\W \y_n$. We observe that the output of cPCA++ is far less noisy than that of the other methods. The number of components was chosen to be $K=3$.}
    \label{fig:mnist_denoising_1}
\end{figure}

In this appendix, we explore the use of the cPCA++ method for matrix factorization and image denoising. This denoising example was performed in \cite{CPCA} for the MNIST over grass dataset examined in Sec.~\ref{sec:mnist}. In this exercise, we are given a single foreground image (i.e., an image containing a digit overlayed on top of grass imagery), flattened into a vector $\z_n\in\mathbb{R}^{M\times 1}$, where $M = 784$ and the subscript $n$ is used to indicate that $\z_n$ is noisy. Since $\z_n$ is a foreground image, it follows the distribution from \eqref{eq:Z1}, and thus we may seek the factorization of the expected value $\E[\z_n] = \W \y_n$ where $\W\in\mathbb{R}^{M\times K}$ and $\y_n\in \mathbb{R}^{K\times 1}$, and $K$ is the effective rank of the denoised image. This is a typical setup of matrix factorization methods. In our case, the matrix $\W$ contains dictionary \emph{atoms} along its columns. These contain the underlying bases that make up the labeled foreground dataset $\Z_f$. On the other hand, the vector $\y_n$ contains the weighting of each dictionary atom (i.e., how much each dictionary atom contributes to the final image) and is a function of the filters $\F$ and the original foreground image $\z_n$. Traditionally, it is assumed that the weighting vector is sparse. We do not directly impose this, but when $K$ is chosen to be small, this is implicitly imposed. That is, we are approximating the foreground image (which is corrupted by the grass background in addition to the actual digit) as:
\begin{align}
    \hat{\z}_n \triangleq \sum_{k=1}^K y_{n,k} \w_{k}
\end{align}
where the atom $\w_k$ denotes the $k$-th column of the matrix $\W$, the weighting $y_{n,k}$ denotes the $k$-th element of the vector $\y_n$, and $\hat{\z}_n$ denotes the denoised version of the image $\z_n$.  When $K\ll M$, we expect that $\W\y_n$ will yield a denoised signal as it will only utilize the few strong atoms forming $\W$ (which best describe the bases present in the foreground dataset). We also saw in Sec.~\ref{sec:cPCA++} that the matrices $\F\in\mathbb{R}^{M\times K}$ and $\W\in\mathbb{R}^{M\times K}$ contain the principal eigenvectors of the matrices $\R_b^{-1} \R_f$ and $\R_f \R_b^{-1}$, respectively. In addition, the vector $\y_n$ is obtained via $\y_n=\F^\T \z_n$. This means that the denoised version of the image is given by $\hat{\z}_n = \W \F^\T \z_n$. We now attempt to denoise an image containing the digit $0$ over grass background. This situation is shown in Fig.~\ref{fig:mnist_denoising_1}. Please  note  that  we  use  the  color  white  to  denote  pixels corresponding  to  the  digit $0$. In the top-left plot, we show the original noisy digit. It is relatively difficult to see the digit over the background, so a significant amount of denoising is necessary. In all of the denoising methods, we set the number of components $K=3$. In the top-right plot, we show the denoising achieved by traditional PCA. In the bottom-left plot, we show the denoising achieved by the cPCA algorithm. Finally, in the bottom-right plot, we show the denoising performance of the cPCA++ method, which is achieved as the low-rank approximation $\W \y_n$. We observe that the output of cPCA++ is far less noisy than that of the other methods.

\section*{Acknowledgment}
This material is based on research sponsored by DARPA and Air Force Research Laboratory (AFRL) under agreement number FA8750-16-2-0173. The U.S. Government is authorized to reproduce and distribute reprints for Governmental purposes notwithstanding any copyright notation thereon. The views and conclusions contained herein are those of the authors and should not be interpreted as necessarily representing the official policies or endorsements, either expressed or implied, of DARPA and Air Force Research Laboratory (AFRL) or the U.S. Government. Credits for the use of the CASIA Image Tampering Detection Evaluation Database (CASIA TIDE) v1.0 and v2.0 are given to the National Laboratory of Pattern Recognition, Institute of Automation, Chinese Academy of Science, Corel Image Database and the photographers http://forensics.idealtest.org.

\ifCLASSOPTIONcaptionsoff
  \newpage
\fi

\bibliographystyle{IEEEtran}
\bibliography{./bibtex/bib/IEEEabrv,./bibtex/bib/refs}

\end{document}